\pdfoutput=1

\documentclass[11pt]{article}

\usepackage[preprint]{acl}
\usepackage{xurl}
\usepackage{times}
\usepackage{latexsym}
\usepackage{algorithm}
\usepackage{algpseudocode}
\usepackage{xcolor}
\usepackage{tcolorbox}
\usepackage{multirow}
\usepackage{booktabs}
\usepackage{makecell}
\usepackage{colortbl}
\usepackage{subcaption}

\newcommand{\inc}[1]{\textcolor{green!60!black}{\textbf{+#1}}}
\newcommand{\dec}[1]{\textcolor{red}{\textbf{-#1}}}
\newcommand{\neu}[1]{\textcolor{gray}{\textbf{#1}}}
\usepackage[T1]{fontenc}

\usepackage[utf8]{inputenc}
\usepackage{amsfonts}

\usepackage{microtype}

\usepackage{inconsolata}

\usepackage{graphicx}

\usepackage{enumitem}

\usepackage{amsmath}
\usepackage{multicol}
\usepackage[normalem]{ulem}

\title{\raisebox{-0.1\height}{\includegraphics[height=1.2em]{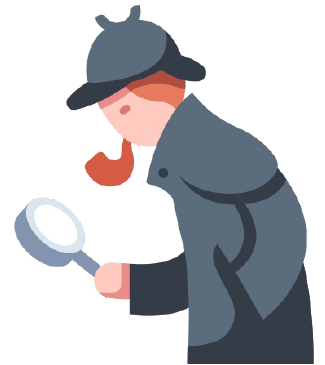}} SpeakerSleuth: Can Large Audio-Language Models \\ Judge Speaker Consistency across Multi-turn Dialogues?}

\author{Jonggeun Lee, Junseong Pyo\textsuperscript{\S}, Gyuhyeon Seo, Yohan Jo$^{\dag}$ \\
Graduate School of Data Science, Seoul National University \\
\texttt{\{jonggeun.lee, yohan.jo\}@snu.ac.kr}}

\begin{document}
\maketitle
\def\thefootnote{\fnsymbol{footnote}}
\footnotetext[2]{Corresponding author.}
\footnotetext[4]{Visiting intern from Hanyang University.}
\def\thefootnote{\arabic{footnote}}
\begin{abstract}
Large Audio-Language Models (LALMs) as judges have emerged as a prominent approach for evaluating speech generation quality, yet their ability to assess speaker consistency across multi-turn dialogues remains unexplored. We present \textbf{SpeakerSleuth}, a benchmark evaluating whether LALMs can reliably judge speaker consistency across multi-turn dialogues through three tasks reflecting real-world requirements.
We construct 1,818 human-verified evaluation instances across four diverse datasets spanning synthetic and real speech, with controlled acoustic difficulty. 
Evaluating twelve widely-used LALMs, we find that models struggle to reliably detect acoustic inconsistencies. For instance, given audio samples of the same speaker's turns, some models overpredict inconsistency, whereas others are overly lenient. Models further struggle to identify the exact turns that are problematic. When other interlocutors' turns are provided as textual context, performance degrades dramatically as models prioritize textual coherence over acoustic cues, failing to detect even obvious gender switches for a speaker.
On the other hand, models perform substantially better in comparing and ranking acoustic variants, demonstrating inherent acoustic discrimination capabilities. These findings expose a significant bias in LALMs: they tend to prioritize text over acoustics, revealing fundamental modality imbalances that need to be addressed to build reliable audio-language judges. Our code and data are available at \url{https://github.com/holi-lab/SpeakerSleuth}.
\end{abstract}


\begin{figure}[!t]
    \centering
    \includegraphics[width=0.98\linewidth]{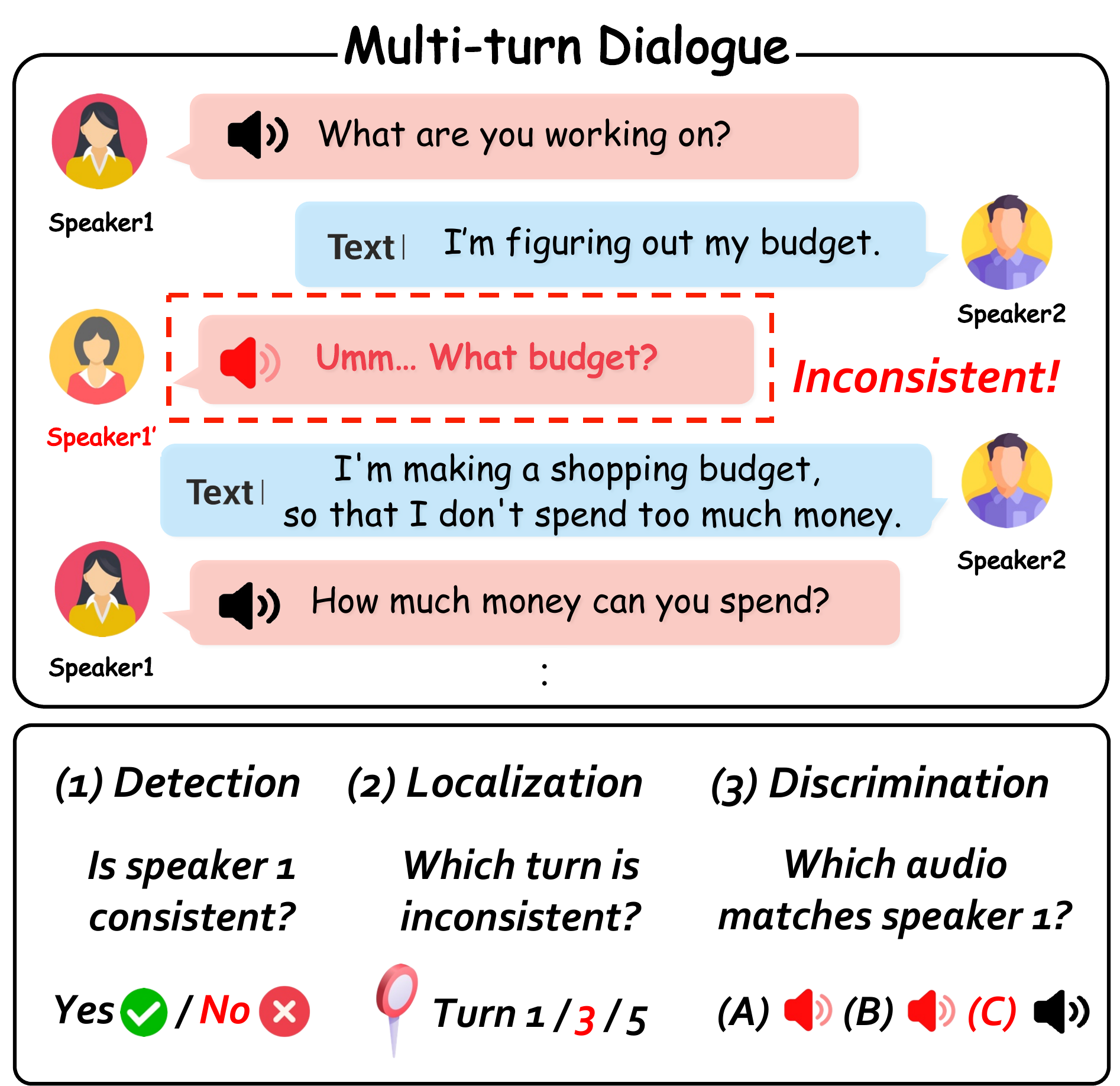}
    \caption{Overview of SpeakerSleuth.}
    \label{fig:intro}
    \vspace{-1em}
\end{figure}
\section{Introduction}

Recent advances in speech synthesis have enabled systems that produce natural, human-like speech \cite{du2024cosyvoice, zhang2024covomix, defossez2024moshi, lee2026spokenus}. These technologies enable diverse applications including voice assistants \cite{apple_intelligence}, voice-overs in podcast generation \cite{notebooklm} and movies \cite{WatchTheSkies2025}, and conversational agents \cite{openai_gpt4o_2024}. A fundamental requirement for these systems is maintaining consistent speaker identity~\cite{Mullennix1990Stimulus}, that is, preserving acoustic characteristics such as timbre, pitch, and voice quality across all utterances in a multi-turn dialogue.
This is particularly important in the speech synthesis of multiple dialogue participants, such as voice-overs in movies.
However, achieving this consistency across long-form, multi-turn dialogues remains challenging~\cite{xie-etal-2025-towards_speech_synthesis}. Even recent models suffer from speaker confusion~\cite{borsos2023audiolm, zhang2024covomix}, timbre drift~\cite{ju2024naturalspeech}, and voice quality variations~\cite{park2025longform}. These failures are particularly difficult to catch because they emerge only 
across turns --- a single generated utterance may sound  natural in isolation, yet be clearly inconsistent when heard in the context of the full dialogue. This 
necessitates reliable verification methods that can assess 
speaker consistency at the dialogue level.

Most approaches~\cite{zhang2024covomix, lee-etal-2025-behavior} evaluate speaker consistency by computing acoustic similarity between utterances using embedding models~\cite{ecapa-tdnn, chen2022wavlm}. However, these methods face fundamental limitations when applied to dialogue evaluation: they operate on pairwise comparisons between two utterances, require manually-set thresholds for binary decisions, and cannot assess consistency holistically across entire dialogues.

Recently, Large Audio-Language Models (LALMs) have emerged as potential alternatives for evaluating speech generation quality \cite{wang2025speechllmasjudgesgeneralinterpretablespeech, wang-etal-2025-qualispeech}. Unlike embedding-based methods that compute pairwise similarities, LALMs can process an entire dialogue at once, receiving both text and audio, and directly outputting a judgment about speaker consistency. 

However, two critical gaps remain: First, no unified benchmark exists to systematically evaluate and compare embedding methods and LALMs for multi-turn speaker consistency assessment. Second, whether LALMs possess the acoustic discrimination capabilities necessary for reliable speaker consistency judgment remains unexplored. 

To address these questions, we present \textbf{SpeakerSleuth}, a benchmark for evaluating both LALMs and embedding-based methods on speaker consistency in multi-turn dialogues. We design our benchmark around three tasks that mirror real-world application requirements (Figure~\ref{fig:intro}). These tasks are \textit{Detection} (identifying whether dialogues contain inconsistencies), \textit{Localization} (pinpointing which specific turns are problematic), and \textit{Discrimination} (comparing and ranking multiple acoustic variants). These capabilities are essential for practical speech generation systems. When dialogue speech is generated, systems must first detect any inconsistencies for each speaker, then localize problematic turns for targeted correction, and finally select optimal outputs from regenerated alternatives. We construct our benchmark from four diverse datasets spanning synthetic and real speech across various conversational settings, comprising 1,818 evaluation instances from 197 speakers, all verified through human annotation.

Our evaluation of 12 state-of-the-art LALMs and 6 embedding methods reveals critical insights into their capabilities. We find that models struggle to reliably detect acoustic inconsistencies due to unstable internal thresholds. This leads to inconsistent decisions where some models are too strict while others are too lenient. Moreover, they struggle with fine-grained turn-level acoustic analysis, as evidenced by their inability to localize specific problematic turns even when detecting overall inconsistency. When other interlocutors' turns are available as dialogue context, LALMs overwhelmingly tend to prioritize textual coherence over acoustic features. They fail to detect even obvious inconsistencies like gender switches for a speaker within coherent dialogue. In contrast, embedding methods achieve stronger detection performance, although they exhibit consistent model-specific biases.

Our contributions are summarized as follows:
\begin{itemize}
\item We present SpeakerSleuth, the first benchmark for multi-turn speaker consistency evaluation with 1,818 human-verified instances.
\item We comprehensively evaluate 12 LALMs and 6 embedding methods, revealing that models struggle with detection and localization.
\item We identify modality imbalances where LALMs prioritize textual context over acoustic discrimination capabilities, providing insights toward reliable audio-language judges.
\end{itemize}

\section{Related Work}

\subsection{Speech Synthesis}
Speech synthesis has evolved from early end-to-end systems~\cite{wang2017tacotron} to sophisticated controllable generation approaches. 
Speaker cloning methods~\cite{wang2023valle,li2025styletts} enabled generating speech in a target 
speaker's voice from reference audio, but lacked intuitive control mechanisms.
Recent text instruction-guided models~\cite{Guo2023PromptttsCT, du2024cosyvoice} allow natural language 
control, significantly improving usability for multi-speaker dialogue 
generation~\cite{zhang2024covomix, zhang2025covomix2, lee-etal-2025-behavior}.

However, maintaining consistent speaker identity across long-form, multi-turn 
dialogues remains challenging~\cite{xie-etal-2025-towards_speech_synthesis}. Models exhibit speaker confusion~\cite{borsos2023audiolm, zhang2024covomix}, timbre drift~\cite{ju2024naturalspeech}, 
and voice quality variations~\cite{park2025longform}, particularly in zero-shot settings 
where limited reference audio must generalize to extended conversations.
These challenges necessitate robust evaluation methods that can reliably assess both speech quality and speaker consistency across multi-turn dialogues.

\subsection{Speech Quality Evaluation}
Speech quality evaluation systematically assesses speech across dimensions such as naturalness and intelligibility~\cite{speechevaluation}.
Traditional approaches rely on objective metrics such as Mel-Cepstral 
Distortion (MCD)~\cite{melcepstral} and PESQ~\cite{pesq}, or subjective human 
assessment through Mean Opinion Score (MOS) \cite{human_mos}. While neural approaches have enabled automated MOS prediction~\cite{saeki2022utmos}, they typically focus on single quality dimensions.
More recently, the LALM-as-a-Judge paradigm~\cite{wang2025speechllmasjudgesgeneralinterpretablespeech, wang-etal-2025-qualispeech, chenaudio_speechjudge, wang2025enabling_speechjudge} has emerged, leveraging LALMs trained on joint audio-text data for multi-dimensional quality analysis with natural language reasoning.
This enables LALMs to potentially integrate acoustic features with conversational context, making them promising candidates for evaluating speaker consistency in dialogue settings.

\subsection{Speaker Consistency Evaluation}
Beyond assessing individual utterance quality, speaker consistency evaluation measures whether a speaker's identity remains stable across multiple utterances in a dialogue.
Existing approaches include embedding-based methods~\cite{all-pair, clustering} using speaker verification models~\cite{ecapa-tdnn, chen2022wavlm} to compute similarity scores across dialogue turns~\cite{zhang2024covomix, zhang2025covomix2, ju2025mooncast, lee-etal-2025-behavior}, and human evaluation~\cite{zhang2024covomix}, which incurs high costs. Given the recent success of LALMs in speech quality evaluation, a natural question arises: can they reliably assess speaker consistency? Their acoustic perception capabilities for this task remain unexplored.


\section{Task Formulation}
\label{sec:task_formulation}
To thoroughly evaluate whether LALMs can reliably distinguish speakers based on acoustic features, we propose an evaluation framework. Rather than relying on a single metric, we decompose speaker consistency into three capabilities: \textit{Detection} (identifying whether all turns are consistent), \textit{Localization} (pinpointing which turn is inconsistent), and \textit{Discrimination} (comparing and ranking acoustic variants by 
their similarity to a target speaker).

Formally, we define a multi-turn dialogue as $\mathcal{D} = \{(t_1, a_1), (t_2, a_2), \ldots, (t_N, a_N)\}$, where $t_i$ represents the transcript and $a_i$ denotes the audio waveform of the $i$-th turn.
Let $I \subset \{1, \ldots, N\}$ denote the indices of turns belonging to a specific target speaker $S$. We denote the audio turns of the target speaker as $\mathcal{A}_S = \{a_i\}_{i \in I}$. In our primary evaluation, models receive $\mathcal{A}_S$ and a reference audio sample of the target speaker to isolate acoustic features from textual cues, with the effect of adding textual context examined separately in Section~\ref{sec:analysis}.

\subsection{Task 1: Detection}
In Text-to-Speech (TTS) and voice cloning, ensuring speaker consistency across generated outputs is critical for quality control. The most fundamental requirement is to detect whether all audio turns belong to the same speaker. This requires \textit{absolute judgment capability}, where the model must rely on its internal threshold to determine consistency.

Given $\mathcal{A}_S$, the model predicts whether all audio turns maintain the identity of speaker $S$ (\texttt{Consistent}) or not (\texttt{Inconsistent}). Success on this task demonstrates that the model possesses an appropriate internal threshold to reliably judge speaker identity based on acoustic features. The prompts are provided in 
Figures~\ref{fig:prompt_detection_audio_only} 
and~\ref{fig:prompt_detection_with_text}.

\subsection{Task 2: Localization}
When speaker inconsistencies are detected in a multi-turn dialogue, identifying the exact problematic turn is essential for efficient correction and regeneration. Merely detecting an anomaly is insufficient; a robust judge must pinpoint where the inconsistency occurs to enable targeted fixes.

Given $\mathcal{A}_S$, the model identifies which turn disrupts speaker identity, or predicts \texttt{None} if all turns are consistent. Success on this task demonstrates that the model can distinguish acoustic speaker characteristics at a fine-grained level, rather than relying on dialogue-level patterns. The prompts are provided in 
Figures~\ref{fig:prompt_localization_audio_only} 
and~\ref{fig:prompt_localization_with_text}.

\begin{figure*}[!t]
    \centering
    \includegraphics[width=1\textwidth]{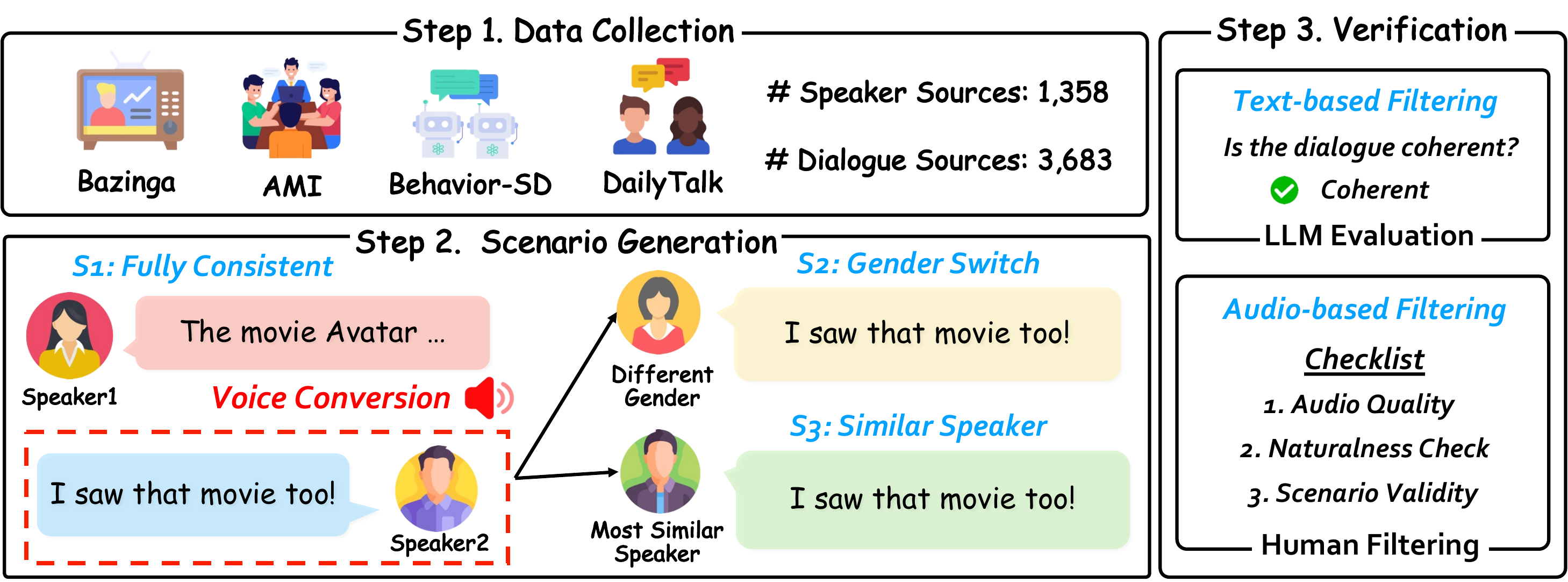}
    \caption{SpeakerSleuth Construction Pipeline.}
    \vspace{-1em}
    \label{fig_overview}
\end{figure*}

\subsection{Task 3: Discrimination}
When an inconsistent turn is identified (Task 2), TTS systems typically regenerate multiple candidate outputs and must select the one that best matches the target speaker. This requires \textit{relative judgment capability}, the ability to compare and rank audio samples by their acoustic similarity to a reference speaker, rather than making absolute binary decisions as in Task 1.

Given $\mathcal{A}_S$ where a target turn is masked, the model is 
presented with three candidates $\mathcal{C} = \{c_1, c_2, c_3\}$ 
representing varying levels of acoustic similarity to the original 
speaker. The order of candidates is randomized to avoid positional 
bias~\cite{pezeshkpour2024mcq_positional_bias}. We evaluate two 
formulations: \textbf{classification}, where the model selects the 
best-matching candidate, and \textbf{ranking}, where the model 
orders all three candidates by acoustic similarity, which poses a strictly harder objective. Success on this task indicates that 
the model can discriminate speakers acoustically. The prompts 
for both formulations are provided in 
Figures~\ref{fig:prompt_discrimination_classification} and~\ref{fig:prompt_discrimination_ranking}.


\subsection{Distinction from Traditional Speaker Recognition}
It is important to note that our tasks differ fundamentally from traditional speaker recognition \cite{furui1996overview, gish1994text}, which consists of two main tasks: Speaker Identification determines \textit{who is speaking} from known speakers, while Speaker Diarization segments continuous audio streams to determine \textit{who spoke when}. In contrast, we evaluate \textit{speaker consistency}: given turns that are assumed or claimed to belong to the same speaker, we assess whether the model can verify that they actually maintain acoustic coherence.

\section{SpeakerSleuth}\label{sec:speakersleuth}

To systematically evaluate the capabilities defined in Section \ref{sec:task_formulation}, we introduce SpeakerSleuth, a benchmark composed of multi-turn dialogues with rigorous acoustic and contextual controls. Figure~\ref{fig_overview} illustrates our benchmark construction pipeline.

\subsection{Step 1. Data Collection}
We collect dialogues with audio and transcripts from four datasets to ensure diversity in conversational domains and styles (Figure~\ref{fig_overview}-1):
\textit{Bazinga}~\cite{lerner-etal-2022-bazinga} contains multi-party dialogues from TV shows and movies (e.g., Friends), testing the model's ability to track speakers in dynamic, scripted interactions. \textit{AMI}~\cite{carletta2005ami} consists of spontaneous business meetings, assessing performance in formal, overlapping, and noisy environments. \textit{Behavior-SD}~\cite{lee-etal-2025-behavior} provides synthesized dialogues with controlled speech behaviors (e.g., fillers, backchannels), enabling evaluation of generated speech. \textit{DailyTalk}~\cite{lee2023dailytalk} captures high-quality everyday conversations, serving as a baseline for casual social interaction.

From these sources, we construct an initial pool of 3,683 dialogues spanning 1,358 unique speakers.

\subsection{Step 2. Scenario Generation}\label{sec:speakersleuth:scenario_generation}
\paragraph{Dialogue Extraction.}
From the collected pool, we select dialogues and extract segments 
where a target speaker appears multiple times across the conversation. 
Each segment contains exactly 5 target speaker turns, with the total number 
of turns across all speakers capped at 20. This cap is necessary because in 
multi-party dialogues, the target speaker's turns are interspersed with 
those of other participants; without the constraint, the gap between target 
turns could grow excessively large. We also sample a reference audio
of the target speaker from outside these segments.

\paragraph{Scenario Generation.}
As illustrated in Figure~\ref{fig_overview}-2, we create three scenarios per dialogue to systematically test acoustic discrimination capabilities: Original conversations serve as positive samples (\textbf{S1: Fully Consistent}), establishing baseline performance when acoustic and textual cues are naturally aligned. To create inconsistent scenarios, we randomly select one turn and apply voice conversion, transforming acoustic timbre while preserving linguistic and prosodic content. For \textbf{S2 (Gender Switch)}, we convert the turn to an opposite-gender voice sampled from the pool, creating clear acoustic deviations. For \textbf{S3 (Similar Speaker)}, we convert the turn to an acoustically similar speaker, selected by computing ECAPA-TDNN~\cite{ecapa-tdnn} embeddings and choosing the one with highest cosine similarity (excluding the target speaker). This requires fine-grained discrimination of subtle timbre differences. By using identical dialogue content across all scenarios, we isolate acoustic features from confounding factors. 

For the primary benchmark, we use FreeVC~\cite{li2023freevc} for voice conversion. We also construct an extended benchmark with CosyVoice3~\cite{du2025cosyvoice3}, OpenVoice~\cite{qin2023openvoice}, and YourTTS~\cite{casanova2022yourtts} to verify robustness across voice conversion models (Appendix~\ref{app:extended_vc}). 

\begin{table}[t]
\centering
\small
\setlength{\tabcolsep}{2pt}
\begin{tabular}{lcccc}
\toprule
Dataset & Instances & Speakers & Avg Turns\textsuperscript{†} & Total Duration \\
\midrule
Bazinga & 636 & 109 & 9.9 ± 1.7 & 3.7 hrs \\
AMI & 138 & 34 & 7.9 ± 3.4 & 0.9 hrs \\
Behavior-SD & 477 & 52 & 7.9 ± 1.2 & 3.0 hrs \\
DailyTalk & 567 & 2 & 9.0 ± 0.0 & 2.6 hrs \\
\midrule
\textbf{Total} & \textbf{1,818} & \textbf{197} & \textbf{8.9 ± 1.7} & \textbf{10.2 hrs} \\
\bottomrule
\multicolumn{5}{l}{\footnotesize \textsuperscript{†}Target speaker: 5 audio turns; other speakers: text.}
\end{tabular}
\caption{Statistics of SpeakerSleuth.}
\label{tab:benchmark_stats}
\vspace{-1em}
\end{table}

\begin{figure}[t]
\centering
\includegraphics[width=1\columnwidth]{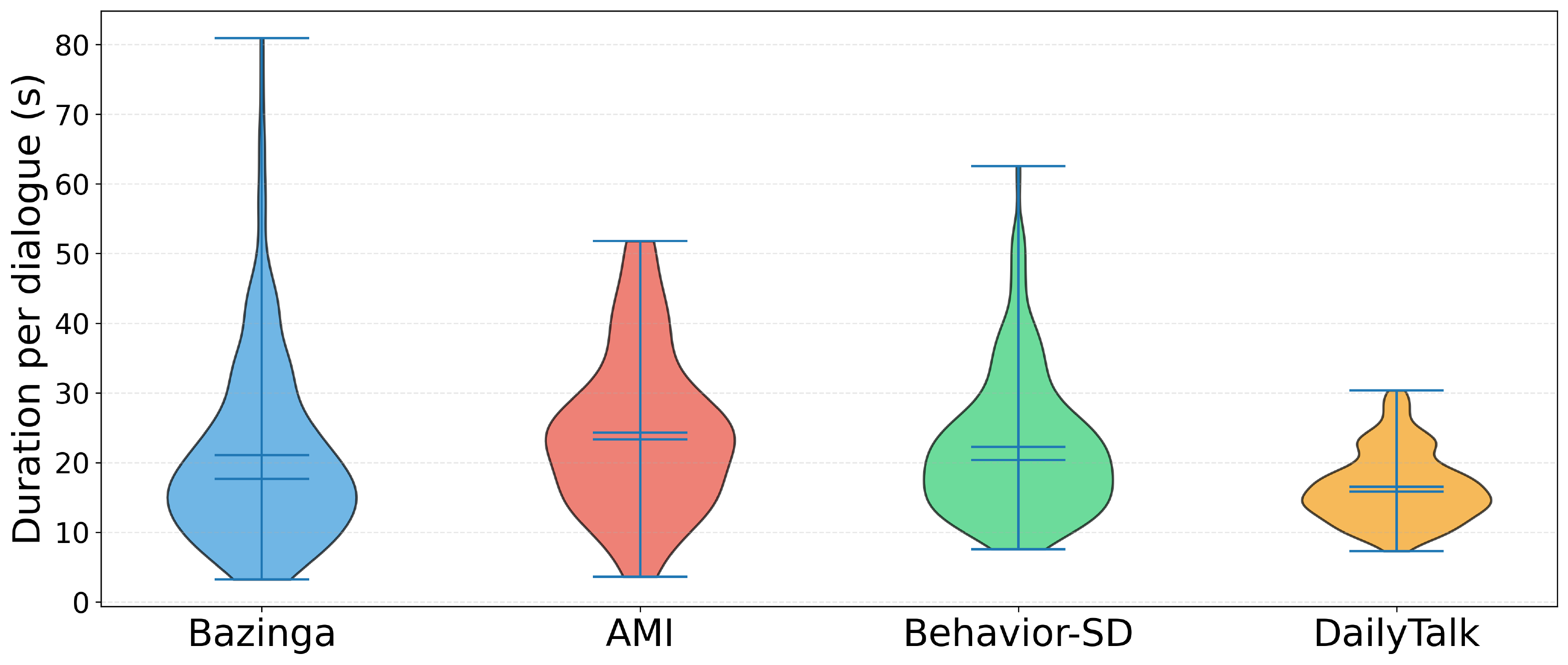}
\caption{Distribution of per-instance audio duration.}
\label{fig:duration_dist}
\vspace{-1.0em}
\end{figure}

\subsection{Step 3. Verification}\label{verification}
As illustrated in Figure~\ref{fig_overview}-3, we validate dialogue 
segments through automated text-based filtering and manual 
audio-based verification.
For text-based filtering, we use Qwen3-32B~\cite{yang2025qwen3technicalreport} 
to filter out segments that lack natural conversational flow 
when isolated from their original 
context~\cite{zhang2024comprehensive}. 
For audio-based verification, expert annotators verify audio 
quality (clarity, absence of noise or artifacts) and naturalness 
of voice-converted turns. Annotators also confirm that each 
scenario exhibits its intended acoustic characteristics. 
Only instances meeting all criteria are retained. 
Details are provided in Appendix~\ref{app:verification_details}.

\begin{figure}[t]
    \centering
    \includegraphics[width=1.0\linewidth]{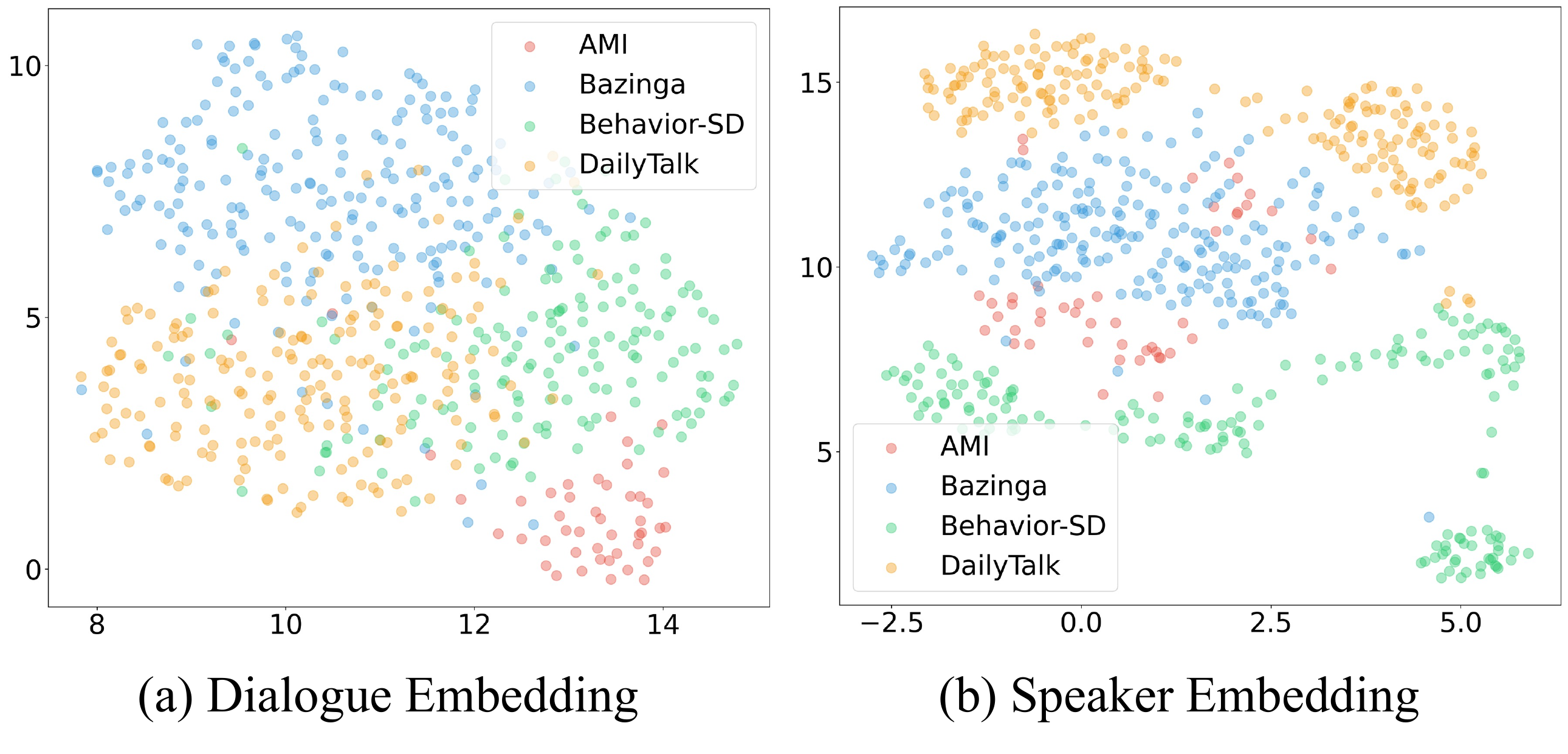}
    \caption{UMAP visualization of dialogue and speaker embeddings.}
    \label{fig:speaker_embedding}
    \vspace{-1.0em}
\end{figure}

\subsection{Benchmark Composition}
Our final benchmark contains 606 unique dialogues, each contributing three scenarios, yielding 1,818 total evaluation instances. The benchmark comprises 10.2 hours of audio from 197 speakers across the four source datasets. Table~\ref{tab:benchmark_stats} summarizes the dataset statistics, and Figure~\ref{fig:duration_dist} shows the distribution of sample durations.
By holding dialogue content constant across scenarios, performance differences between S1, S2, and S3 directly reflect the model's ability to detect varying degrees of acoustic deviation. UMAP~\cite{mcinnes2018umap} visualization of dialogue and speaker embeddings confirms substantial diversity in both dialogue content and acoustic characteristics (Figure~\ref{fig:speaker_embedding}; details in 
Appendix~\ref{app:embedding_details}).

\begin{table*}[!ht]
\centering
\scriptsize
\setlength{\tabcolsep}{3pt}
\begin{tabular}{l|cccc|c|ccc|ccc|c|ccccc|ccc}
\toprule
& \multicolumn{4}{c|}{\textbf{Detection}} & \multicolumn{8}{c|}{\textbf{Localization}} & \multicolumn{8}{c}{\textbf{Discrimination}} \\
\cmidrule(lr){2-5} \cmidrule(lr){6-13} \cmidrule(lr){14-21}
\textbf{Model / Method} & S1 & S2 & S3 & Bal & \multicolumn{1}{c|}{S1} & \multicolumn{3}{c|}{S2} & \multicolumn{3}{c|}{S3} & Bal & \multicolumn{5}{c|}{Classification} & \multicolumn{3}{c}{Ranking} \\
\cmidrule(lr){2-5} \cmidrule(lr){6-6} \cmidrule(lr){7-9} \cmidrule(lr){10-12} \cmidrule(lr){13-13} \cmidrule(lr){14-18} \cmidrule(lr){19-21}
& \multicolumn{4}{c|}{Acc} & F1 & P & R & F1 & P & R & F1 & F1 & Baz & AMI & B-SD & Daily & Avg & N@1 & N@2 & EM \\
\midrule
\rowcolor{yellow!30}
\multicolumn{21}{c}{\textit{Large Audio-Language Models}} \\
GPT-4o-audio & 72.9 & 32.8 & 29.5 & 52.0 & 71.5 & 12.6 & 25.9 & 15.0 & 8.7 & 19.2 & 11.1 & 42.3 & 50.2 & 30.4 & 38.4 & 34.4 & 40.7 & 45.4 & 60.1 & 19.8 \\
Gemini-2.5-Pro & 73.9 & 71.6 & 39.3 & \textbf{64.7} & 65.2 & 59.3 & 70.3 & 62.5 & 44.0 & 56.6 & 47.5 & \textbf{60.1} & 77.8 & 76.1 & 81.1 & 87.2 & \textbf{81.5} & \textbf{88.8} & \textbf{92.6} & \textbf{71.5} \\
Gemini-2.5-Flash & 97.4 & 45.5 & 12.0 & \underline{63.1} & 64.4 & 55.6 & 90.1 & 62.3 & 38.6 & 72.4 & 45.0 & \underline{59.0} & 70.8 & 73.9 & 69.2 & 86.8 & \underline{75.6} & \underline{83.6} & \underline{88.3} & \underline{61.6} \\
Gemini-2.5-Flash-Lite & 40.8 & 70.3 & 69.3 & 55.3 & 0.3 & 25.8 & 94.7 & 36.4 & 24.5 & 94.1 & 35.4 & 18.1 & 47.6 & 45.7 & 47.8 & 44.4 & 46.5 & 53.5 & 64.6 & 23.1 \\
Qwen2.5-Omni-3B & 54.5 & 49.3 & 47.8 & 51.5 & 21.8 & 45.6 & 91.6 & 47.7 & 28.9 & 81.6 & 36.0 & 31.8 & 42.9 & 39.1 & 41.5 & 38.1 & 40.8 & 51.7 & 63.1 & 22.8 \\
Qwen2.5-Omni-7B & 32.0 & 72.3 & 69.8 & 51.5 & 38.0 & 14.1 & 61.8 & 22.6 & 13.2 & 57.5 & 21.0 & 29.9 & 57.5 & 30.4 & 39.6 & 31.7 & 42.7 & 52.4 & 64.2 & 22.9 \\
Qwen3-Omni-30B-A3B & 88.1 & 29.9 & 14.0 & 55.0 & 0.2 & 15.1 & 57.4 & 23.3 & 15.2 & 56.0 & 23.5 & 11.8 & 82.1 & 69.6 & 43.4 & 48.1 & 60.4 & 66.9 & 73.2 & 36.8 \\
MiniCPM-o-2.6 & 85.3 & 0.7 & 0.0 & 42.8 & 66.9 & 29.8 & 54.5 & 35.4 & 12.3 & 25.1 & 17.2 & 46.6 & 62.7 & 45.7 & 42.8 & 41.3 & 49.5 & 52.2 & 65.4 & 23.6 \\
Gemma-3n-E4B & 51.2 & 50.6 & 48.4 & 50.4 & 0.0 & 19.1 & 95.1 & 32.0 & 19.1 & 94.4 & 32.0 & 16.0 & 36.8 & 37.0 & 33.3 & 40.7 & 37.1 & 47.5 & 62.1 & 20.1 \\
Phi-4-multimodal & 78.8 & 24.9 & 24.3 & 51.7 & 99.7 & 0.0 & 0.1 & 0.1 & 0.0 & 0.1 & 0.1 & 49.9 & 37.3 & 39.1 & 39.0 & 43.9 & 39.9 & 49.5 & 63.3 & 21.3 \\
Omnivinci & 81.6 & 46.1 & 20.9 & 57.5 & 48.3 & 14.9 & 61.2 & 23.6 & 8.8 & 40.0 & 15.2 & 33.8 & 51.4 & 39.1 & 42.1 & 42.9 & 45.4 & 57.0 & 67.3 & 25.4 \\
Audio-Flamingo-3 & 99.2 & 1.3 & 1.2 & 50.2 & 0.0 & 17.1 & 38.4 & 23.3 & 17.4 & 38.8 & 23.7 & 11.8 & 33.0 & 28.3 & 37.1 & 41.3 & 36.3 & 44.9 & 60.7 & 18.3 \\
\midrule
\rowcolor{blue!20}
\multicolumn{21}{c}{\textit{Speaker Embedding Methods}} \\
Pairwise (ECAPA) & 36.0 & 88.4 & 86.3 & 61.7 & 36.0 & 44.1 & 92.2 & 46.5 & 38.3 & 81.8 & 43.5 & 40.5 & 97.8 & 99.0 & 100.0 & 100.0 & \textbf{99.2} & \textbf{99.6} & \textbf{92.7} & \textbf{58.6} \\
Pairwise (WavLM) & 91.8 & 38.4 & 37.7 & \textbf{64.9} & 91.8 & 36.0 & 37.8 & 31.8 & 36.8 & 38.5 & 31.3 & \textbf{61.7} & 96.7 & 95.5 & 79.9 & 100.0 & \underline{93.2} & \underline{94.4} & \underline{91.0} & \underline{55.0} \\
Centroid (ECAPA) & 47.3 & 75.7 & 73.7 & 61.0 & 47.3 & 54.1 & 90.7 & 46.3 & 47.9 & 80.8 & 43.3 & 46.0 & 97.8 & 99.0 & 100.0 & 100.0 & \textbf{99.2} & \textbf{99.6} & \textbf{92.7} & \textbf{58.6} \\
Centroid (WavLM) & 87.5 & 38.2 & 37.7 & \underline{62.7} & 87.5 & 49.0 & 51.3 & 31.6 & 52.1 & 54.3 & 31.0 & \underline{59.4} & 96.7 & 95.5 & 79.9 & 100.0 & \underline{93.2} & \underline{94.4} & \underline{91.0} & \underline{55.0} \\
Reference (ECAPA) & 8.5 & 95.5 & 94.5 & 51.8 & 8.5 & 33.7 & 89.8 & 39.0 & 29.3 & 81.2 & 34.6 & 22.6 & 97.8 & 92.0 & 99.4 & 76.2 & 91.0 & 92.3 & 87.2 & 52.7 \\
Reference (WavLM) & 79.3 & 32.2 & 32.6 & 55.9 & 79.3 & 23.8 & 25.7 & 16.9 & 23.0 & 24.6 & 17.1 & 48.3 & 100.0 & 94.3 & 73.0 & 68.8 & 82.8 & 88.3 & 84.3 
& 51.4 \\
\bottomrule
\end{tabular}
\caption{Main results (\%). Detection reports per-scenario and Balanced Accuracy (Bal). Localization reports F1 for S1, Precision (P), Recall (R), and F1 for S2/S3, and Balanced F1. Discrimination reports per-dataset and average Classification Accuracy, and Ranking metrics (N@1: NDCG@1, N@2: NDCG@2, EM: Exact Match). \textbf{Bold} indicates the highest and \underline{underline} the second-highest per group for Bal Acc, Bal F1, Avg, N@1, N@2, and EM.}
\label{tab:main_results}
\end{table*}

\section{Experimental Setup}
\subsection{Models}
\paragraph{LALM Judges.} 
We assess twelve widely-used LALMs: GPT-4o-audio 
\cite{openai_gpt4o_2024}, Gemini-2.5-Pro \cite{google_gemini25pro_2025}, 
Gemini-2.5-Flash/Flash-Lite~\cite{google_gemini25flash_2025}, 
Qwen2.5-Omni-3B/7B \cite{xu2025qwen25omnitechnicalreport}, 
Qwen3-Omni-30B-A3B \cite{xu2025qwen3omnitechnicalreport}, 
MiniCPM-o-2.6 \cite{minicpm-o-2.6}, 
Gemma-3n-E4B \cite{gemmateam2025gemma3technicalreport}, 
Phi-4-multimodal \cite{microsoft2025phi4minitechnicalreportcompact}, 
Omnivinci \cite{ye2025omnivinci}, and 
Audio-Flamingo-3 \cite{ghosh2025audioflamingo3}. 
These models vary in architecture and scale, enabling analysis 
of how model capacity affects consistency evaluation.

\paragraph{Speaker Embedding Methods.}\label{sec:speaker_embedding}
We evaluate three speaker embedding methods, each with two backbone models: WavLM~\cite{chen2022wavlm} and ECAPA-TDNN~\cite{ecapa-tdnn}. All methods flag turns as inconsistent when their similarity (or equivalently distance) crosses a threshold $\tau$, flag all such turns for localization, and rank candidates via their similarity metric for discrimination.
\textbf{Pairwise Similarity} computes each turn's average 
cosine similarity to all other turns ($\tau_{\text{pair}} = 0.4$). 
\textbf{Centroid Distance} computes each turn's distance from 
the centroid of all turns ($\tau_{\text{cent}} = 0.3$). \textbf{Reference Comparison} computes similarity between each embedding and a reference speaker audio ($\tau_{\text{ref}} = 0.4$).
Thresholds are set based on preliminary validation. Detailed algorithms and threshold sensitivity analysis are in Appendix~\ref{embedding_methods_details}.

\subsection{Evaluation Protocol}\label{evaluation_protocol}
As described in Section~\ref{sec:task_formulation}, LALMs receive all target speaker turns at once and judge consistency across the full dialogue. Our primary evaluation uses audio-only input with a reference sample of at least 3 seconds~\cite{wang2023valle} from the target speaker. This setting reflects realistic TTS scenarios where target speaker samples guide generation~\cite{li2025styletts}, and enables fair comparison with embedding-based methods that also operate on audio-only input. To isolate the contribution of different factors, we also examine per-turn comparison against the reference (same as Reference comparison in Speaker Embedding Methods; full results in Appendix~\ref{app:pairwise}), removal of the reference audio, and addition of textual context (Section~\ref{sec:analysis}).

\paragraph{Evaluation Metrics.}
For Detection, we report per-scenario accuracy and Balanced Accuracy:
\[
\text{Bal} = \frac{1}{2}\left(\text{Acc}_{S1} + \frac{\text{Acc}_{S2} + \text{Acc}_{S3}}{2}\right)
\]
which equally weights consistent (S1) and inconsistent (S2/S3) scenarios. This balancing is necessary because S1 and S2/S3 penalize opposite model behaviors: a model that always predicts \texttt{consistent} achieves 100\% on S1 but 0\% on S2/S3, and vice versa.
For Localization, we compute Precision, Recall, and F1-score
per dialogue instance and report their macro-averages across
all instances for each scenario, along with Balanced F1 following the 
same balancing scheme. 
For Discrimination, the 
classification reports per-dataset accuracy and overall 
accuracy across all samples, while the ranking reports 
NDCG@1, NDCG@2, and Exact Match. 
Detailed definitions are in 
Appendix~\ref{evaluation_metrics}.

\section{Main Results}\label{sec:main_results}
\subsection{Detection and Localization Performance}
\begin{figure}[t]
\centering
\hspace{-1em}
\includegraphics[width=0.95\columnwidth]{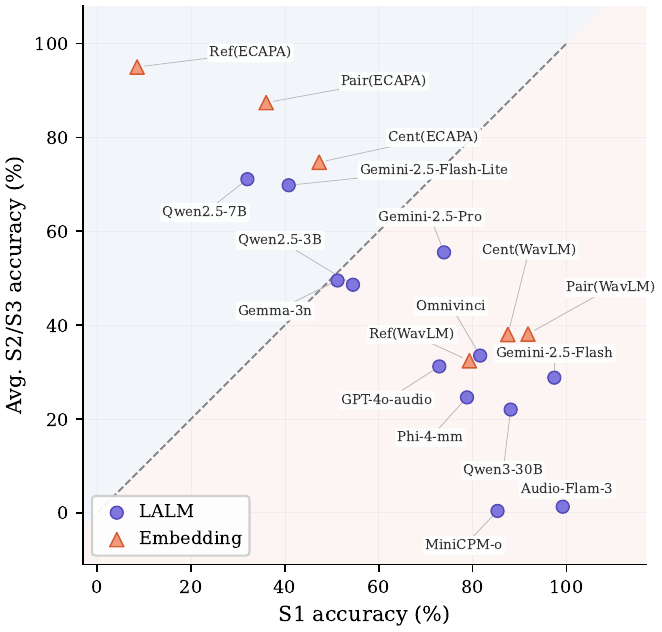}
\caption{\textbf{Detection threshold bias across models.} Each point shows a model's S1 accuracy vs.\ average S2/S3 accuracy.}
\label{fig:detection_scatter}
\vspace{-1em}
\end{figure}

\paragraph{LALMs.}
As shown in Table~\ref{tab:main_results} and 
Figure~\ref{fig:detection_scatter}, detection results reveal 
that LALMs lack balanced internal thresholds for speaker 
consistency judgment. Models cluster along the anti-diagonal (i.e., high S1 with low S2/S3, or vice versa): 
those like MiniCPM-o-2.6 and Audio-Flamingo-3 overwhelmingly 
predict \texttt{consistent}, failing to detect even obvious speaker 
changes, while Gemini-2.5-Flash-Lite and Qwen2.5-Omni-7B 
exhibit the opposite bias. This instability results in poor 
balanced accuracy, with most models scoring below 60\%. The 
best model, Gemini-2.5-Pro, achieves 64.7\% but remains 
notably weak on S3 (39.3\%), indicating it can detect gender 
switches but struggles when the substituted speaker is 
acoustically similar.

Localization results further reveal the limitations of current 
LALMs. Most models exhibit extreme behavior: some default to 
marking no turns as inconsistent (e.g., Phi-4-multimodal with 
near-zero F1 across S2/S3), while others flag nearly all turns 
indiscriminately (e.g., Gemma-3n-E4B with ~95\% recall but 
~19\% precision, at the chance level for 5 turns). Critically, models in the latter group show near-identical scores across S2 and S3, confirming that they 
cannot distinguish gender switches from subtle timbre 
differences. These models flag turns based on a fixed bias rather than 
actual acoustic content. Only Gemini-2.5-Pro (S2/S3 F1: 
62.5\%/47.5\%) and Gemini-2.5-Flash (62.3\%/45.0\%) maintain 
meaningful precision alongside high recall, and notably show 
a drop from S2 to S3, indicating that these models do respond 
to acoustic difficulty.

\paragraph{Speaker Embedding Methods.}
Speaker embedding methods achieve comparable detection performance, 
with Pairwise (WavLM) reaching 64.9\% balanced accuracy, but 
exhibit the same systematic biases: ECAPA-TDNN-based methods 
over-detect changes while under-performing on S1, and 
WavLM-based methods show the opposite pattern. For localization, 
even methods with strong detection do not proportionally improve 
at pinpointing the inconsistent turn (best: 61.7\% balanced F1), 
suggesting that the ability to detect inconsistency does not transfer to pinpointing where it occurs.

\begin{table*}[!ht]
\centering
\scriptsize
\setlength{\tabcolsep}{2.8pt}
\begin{tabular}{l|rrr|rrr|rrr||rrr|rrr|rrr}
\toprule
& \multicolumn{9}{c||}{\textbf{Impact of Textual Context}} & \multicolumn{9}{c}{\textbf{Impact of Reference Audio}} \\
\cmidrule(lr){2-10} \cmidrule(lr){11-19}
& \multicolumn{3}{c|}{\textbf{S1}} & \multicolumn{3}{c|}{\textbf{S2}} & \multicolumn{3}{c||}{\textbf{S3}} & \multicolumn{3}{c|}{\textbf{S1}} & \multicolumn{3}{c|}{\textbf{S2}} & \multicolumn{3}{c}{\textbf{S3}} \\
\cmidrule(lr){2-4} \cmidrule(lr){5-7} \cmidrule(lr){8-10} \cmidrule(lr){11-13} \cmidrule(lr){14-16} \cmidrule(lr){17-19}
\textbf{Model} & Audio & +C & $\Delta$ & Audio & +C & $\Delta$ & Audio & +C & $\Delta$ & w/ Ref & w/o & $\Delta$ & w/ Ref & w/o & $\Delta$ & w/ Ref & w/o & $\Delta$ \\
\midrule
GPT-4o-audio & 72.9 & 93.4 & \inc{20.5} & 32.8 & 6.3 & \dec{26.5} & 29.5 & 5.0 & \dec{24.5} & 72.9 & 80.5 & \inc{7.6} & 32.8 & 16.2 & \dec{16.6} & 29.5 & 13.9 & \dec{15.6} \\
Gemini-2.5-Pro & 73.9 & 34.5 & \dec{39.4} & 71.6 & 46.8 & \dec{24.8} & 39.3 & 32.8 & \dec{6.5} & 73.9 & 47.4 & \dec{26.5} & 71.6 & 38.4 & \dec{33.2} & 39.3 & 16.7 & \dec{22.6} \\
Gemini-2.5-Flash & 97.4 & 91.9 & \dec{5.5} & 45.5 & 16.7 & \dec{28.8} & 12.0 & 10.7 & \dec{1.3} & 97.4 & 97.0 & \dec{0.4} & 45.5 & 41.3 & \dec{4.2} & 12.0 & 15.5 & \inc{3.5} \\
Gemini-2.5-Flash-Lite & 40.8 & 93.4 & \inc{52.6} & 70.3 & 3.3 & \dec{67.0} & 69.3 & 3.3 & \dec{66.0} & 40.8 & 92.6 & \inc{51.8} & 70.3 & 13.9 & \dec{56.4} & 69.3 & 12.5 & \dec{56.8} \\
Qwen2.5-Omni-3B & 54.5 & 85.3 & \inc{30.8} & 49.3 & 15.8 & \dec{33.5} & 47.8 & 12.7 & \dec{35.1} & 54.5 & 92.7 & \inc{38.2} & 49.3 & 10.6 & \dec{38.7} & 47.8 & 9.0 & \dec{38.8} \\
Qwen2.5-Omni-7B & 32.0 & 59.8 & \inc{27.8} & 72.3 & 41.7 & \dec{30.6} & 69.8 & 42.5 & \dec{27.3} & 32.0 & 98.5 & \inc{66.5} & 72.3 & 2.8 & \dec{69.5} & 69.8 & 2.0 & \dec{67.8} \\
Qwen3-Omni-30B-A3B & 88.1 & 93.1 & \inc{5.0} & 29.9 & 8.6 & \dec{21.3} & 14.0 & 6.9 & \dec{7.1} & 88.1 & 99.8 & \inc{11.7} & 29.9 & 11.7 & \dec{18.2} & 14.0 & 1.2 & \dec{12.8} \\
MiniCPM-o-2.6 & 85.3 & 85.7 & \inc{0.4} & 0.7 & 0.0 & \dec{0.7} & 0.0 & 0.0 & \neu{0.0} & 85.3 & 86.5 & \inc{1.2} & 0.7 & 3.1 & \inc{2.4} & 0.0 & 0.3 & \inc{0.3} \\
Gemma-3n-E4B & 51.2 & 93.9 & \inc{42.7} & 50.6 & 5.8 & \dec{44.8} & 48.4 & 5.8 & \dec{42.6} & 51.2 & 36.4 & \dec{14.8} & 50.6 & 66.9 & \inc{16.3} & 48.4 & 65.6 & \inc{17.2} \\
Phi-4-multimodal & 78.8 & 61.5 & \dec{17.3} & 24.9 & 39.2 & \inc{14.3} & 24.3 & 38.2 & \inc{13.9} & 78.8 & 92.5 & \inc{13.7} & 24.9 & 8.1 & \dec{16.8} & 24.3 & 7.3 & \dec{17.0} \\
Omnivinci & 81.6 & 98.7 & \inc{17.1} & 46.1 & 2.3 & \dec{43.8} & 20.9 & 1.5 & \dec{19.4} & 81.6 & 88.1 & \inc{6.5} & 46.1 & 61.5 & \inc{15.4} & 20.9 & 16.3 & \dec{4.6} \\
Audio-Flamingo-3 & 99.2 & 97.5 & \dec{1.7} & 1.3 & 1.8 & \inc{0.5} & 1.2 & 1.7 & \inc{0.5} & 99.2 & 95.7 & \dec{3.5} & 1.3 & 4.3 & \inc{3.0} & 1.2 & 3.8 & \inc{2.6} \\
\bottomrule
\end{tabular}
\caption{\textbf{Impact of Textual Context and Reference Audio on Detection Accuracy (\%).} Left: performance change when adding textual context (+C) vs.\ audio-only (Audio). Right: performance change when removing reference audio (w/o) vs.\ with reference (w/ Ref). $\Delta$ denotes the difference.}
\label{tab:analysis_impact}
\vspace{-1em}
\end{table*}

\subsection{Discrimination Performance}
\paragraph{LALMs.}
Table~\ref{tab:main_results} presents discrimination results. 
Compared to detection, discrimination performance improves 
substantially for stronger models: Gemini-2.5-Pro achieves 
81.5\% classification accuracy with 92.6\% NDCG@2 and 71.5\% 
Exact Match, followed by Gemini-2.5-Flash (75.6\%, 88.3\%, 
61.6\%) and Qwen3-Omni-30B-A3B (60.4\%, 73.2\%, 36.8\%). This 
dissociation between detection and discrimination validates 
our task design: models that struggle with absolute binary 
judgments due to unstable thresholds can still perceive acoustic 
differences when comparing candidates.

\paragraph{Speaker Embedding Methods.}
Pairwise and Centroid methods achieve near-perfect classification 
(93--99\%) and high NDCG@2 (91--93\%), but Exact Match drops 
to 55--59\%: they reliably identify the best match but cannot 
order the remaining candidates. Gemini-2.5-Pro shows the 
opposite tradeoff, with lower classification accuracy (81.5\%) 
but substantially higher Exact Match (71.5\%). Embedding 
methods and LALMs thus exhibit complementary strengths on 
discrimination: embeddings excel at pinpointing the closest 
match, while Gemini-2.5-Pro better captures the relative 
ordering among candidates.

\section{Further Analyses of LALMs}
\label{sec:analysis}

\subsection{Impact of Textual Context}\label{sec:analysis_textual} 
Our main results evaluate models using only the target speaker's 
audio turns. A natural hypothesis is that providing the 
interlocutors' turns as text would help models better focus on 
the target speaker's voice by anchoring the conversational flow, 
allowing them to allocate more attention to acoustic features 
of the target. To test this, we provide the full dialogue to 
LALM judges, with non-target interlocutors' turns in text form 
while the target speaker's turns remain as audio. By construction, all dialogues are textually coherent (Section~\ref{verification}), so the text itself offers no signal of inconsistency, allowing us to test whether textual context helps models 
focus on acoustic features.

Table~\ref{tab:analysis_impact} reveals the opposite: adding textual context degrades rather than improves acoustic judgment. For most models, it sharply improves S1 accuracy while collapsing S2/S3 accuracy, with the exception of Gemini-2.5-Pro, which degrades across all scenarios, and Phi-4-multimodal, whose audio-only baseline already collapses, inverting the pattern. This asymmetric 
pattern indicates that models default to judging speakers as 
consistent whenever the dialogue text flows naturally, 
regardless of acoustic evidence, even failing to detect 
obvious gender switches. Rather than helping models focus on the target speaker's voice, 
textual context shifts their judgment toward text-based reasoning, 
revealing a modality imbalance that may reflect disproportionate 
attention to text over audio tokens in LALMs~\cite{wang2025paymoreattentiontoaudio}.
Localization results exhibit a similar pattern 
(Table~\ref{tab:analysis_impact_localization}). These findings reveal that improving LALMs as speaker consistency judges requires not just better acoustic representations, but mechanisms to balance attention allocation during multi-modal fusion. Developing methods to better leverage dialogue context while maintaining acoustic sensitivity would be a promising future direction.

\subsection{Impact of Reference Audio}\label{sec:analysis_reference}

In our main evaluation, we provide models with a single reference audio sample from the target speaker to establish a comparison baseline. We investigate what happens when this reference is absent, requiring models to judge consistency solely from the dialogue turns themselves.

Table~\ref{tab:analysis_impact} shows that for several models, 
removing reference audio leads them to default to 
\textit{Consistent} judgments: S1 accuracy increases sharply 
while S2/S3 accuracy drops substantially. Without an explicit 
comparison anchor, these models adopt lenient thresholds, 
failing to detect even obvious inconsistencies such as gender 
switches. Localization results exhibit the same pattern 
(Table~\ref{tab:analysis_impact_localization}). This reveals that many LALMs fail to establish appropriate decision boundaries without explicit references. 
Rather than developing robust internal speaker representations 
from dialogue context alone, they rely heavily on explicit 
reference audio to anchor their judgments. This pattern 
connects to our earlier finding from the Discrimination task 
(Table~\ref{tab:main_results}), where models achieved 
 better performance through relative judgment. 
These findings have important practical implications: in 
real-world applications such as TTS validation, providing 
reference audio is essential for reliable speaker consistency 
judgment.

\subsection{Effect of Speaker Turns and Clip Duration}\label{sec:10turn}
To examine whether our findings generalize across dialogue 
lengths, we construct a 10-turn dataset comprising 759 instances 
from 253 unique dialogues using the same pipeline as the primary 
benchmark. As shown in Table~\ref{tab:10turn}, the 10-turn 
setting is slightly more challenging on average ($-$1.2\% 
Detection, $-$4.2\% Localization, $-$3.1\% Discrimination), 
though individual trends vary. Model rankings are largely 
preserved, indicating our findings are not specific to the 
5-turn setting (full results in Table~\ref{tab:10turn_full}; analysis in Appendix~\ref{app:10turn}).

\begin{figure}[t]
\centering
\includegraphics[width=\columnwidth]{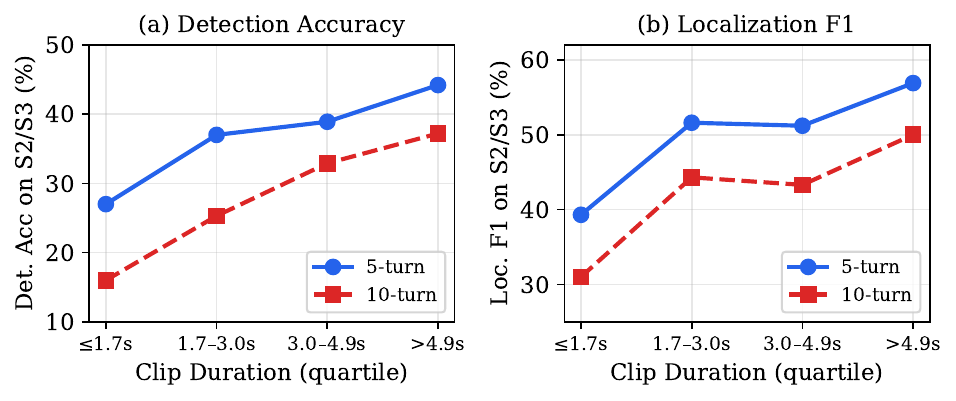}
\caption{Performance on inconsistent scenarios (S2/S3) by clip duration 
quartile for the top three LALMs.}
\label{fig:clip_duration}
\vspace{-1em}
\end{figure}

We further analyze how clip duration affects performance on 
inconsistent scenarios (S2/S3). 
Figure~\ref{fig:clip_duration} shows results for the top three 
LALMs by Discrimination accuracy (Gemini-2.5-Pro, 
Gemini-2.5-Flash, Qwen3-Omni-30B-A3B), grouped by duration quartile. 
Both settings exhibit a clear monotonic trend: longer clips yield higher Detection Accuracy and Localization F1, 
confirming that models require sufficient acoustic evidence 
within each clip for reliable speaker judgment.

\begin{table}[t]
\centering
\scriptsize
\setlength{\tabcolsep}{2pt}
\begin{tabular}{l|ccc|ccc|ccc}
\toprule
& \multicolumn{3}{c|}{\textbf{Det Bal (\%)}} & \multicolumn{3}{c|}{\textbf{Loc Bal F1 (\%)}} & \multicolumn{3}{c}{\textbf{Disc Acc (\%)}} \\
\cmidrule(lr){2-4} \cmidrule(lr){5-7} \cmidrule(lr){8-10}
\textbf{Model} & 5t & 10t & $\Delta$ & 5t & 10t & $\Delta$ & 5t & 10t & $\Delta$ \\
\midrule
GPT-4o-audio & 52.0 & 51.3 & \dec{0.7} & 42.3 & 38.8 & \dec{3.5} & 40.7 & 39.2 & \dec{1.5} \\
Gemini-2.5-Pro & \textbf{64.7} & \textbf{69.3} & \inc{4.6} & \textbf{60.1} & 55.4 & \dec{4.7} & \textbf{81.5} & 80.1 & \dec{1.4} \\
Gemini-2.5-Flash & 63.1 & 56.3 & \dec{6.8} & 59.0 & \textbf{62.4} & \inc{3.4} & 75.6 & \textbf{82.2} & \inc{6.6} \\
Gemini-2.5-Flash-Lite & 55.3 & 48.1 & \dec{7.2} & 18.1 & 15.1 & \dec{3.0} & 46.5 & 46.6 & \inc{0.1} \\
Qwen2.5-Omni-3B & 51.5 & 51.8 & \inc{0.3} & 31.8 & 19.8 & \dec{12.0} & 40.8 & 40.7 & \dec{0.1} \\
Qwen2.5-Omni-7B & 51.5 & 51.1 & \dec{0.4} & 29.9 & 35.6 & \inc{5.7} & 42.7 & 38.7 & \dec{4.0} \\
Qwen3-Omni-30B-A3B & 55.0 & 50.9 & \dec{4.1} & 11.8 & 15.4 & \inc{3.6} & 60.4 & 54.9 & \dec{5.5} \\
MiniCPM-o-2.6 & 42.8 & 50.2 & \inc{7.4} & 46.6 & 36.3 & \dec{10.3} & 49.5 & 37.9 & \dec{11.6} \\
Gemma-3n-E4B & 50.4 & 51.5 & \inc{1.1} & 16.0 & 8.2 & \dec{7.8} & 37.1 & 35.2 & \dec{1.9} \\
Phi-4-multimodal & 51.7 & 50.2 & \dec{1.5} & 49.9 & 50.0 & \inc{0.1} & 39.9 & 31.6 & \dec{8.3} \\
Omnivinci & 57.5 & 52.0 & \dec{5.5} & 33.8 & 16.4 & \dec{17.4} & 45.4 & 40.3 & \dec{5.1} \\
Audio-Flamingo-3 & 50.2 & 48.5 & \dec{1.7} & 11.8 & 7.7 & \dec{4.1} & 36.3 & 32.0 & \dec{4.3} \\
\bottomrule
\end{tabular}
\caption{5-turn vs.\ 10-turn comparison. Det Bal: Balanced Detection Accuracy. Loc Bal F1: Balanced Localization F1. Disc Acc: Discrimination classification accuracy. $\Delta$ = 10t $-$ 5t.}
\label{tab:10turn}
\vspace{-1em}
\end{table}

\begin{table}[t]
\centering
\scriptsize
\setlength{\tabcolsep}{4pt}
\begin{tabular}{l|cccc}
\toprule
\textbf{Model} & Acc & NDCG@1 & NDCG@2 & EM \\
\midrule
GPT-4o-audio & 7.3 & 10.5 & 12.7 & 1.7 \\
Gemini-2.5-Pro & \textbf{48.0} & \textbf{63.0} & \textbf{75.1} & \textbf{16.8} \\
Gemini-2.5-Flash & 37.1 & 53.8 & 67.4 & 8.9 \\
Gemini-2.5-Flash-Lite & 27.0 & 43.8 & 55.2 & 5.6 \\
Qwen2.5-Omni-3B & 23.1 & 39.2 & 51.2 & 4.9 \\
Qwen2.5-Omni-7B & 25.8 & 41.9 & 53.6 & 5.4 \\
Qwen3-Omni-30B-A3B & 37.1 & 52.9 & 62.7 & 11.5 \\
MiniCPM-o-2.6 & 32.0 & 45.4 & 57.1 & 6.6 \\
Gemma-3n-E4B & 26.0 & 40.0 & 49.3 & 3.5 \\
Phi-4-multimodal & 27.6 & 42.2 & 51.3 & 5.4 \\
Omnivinci & 27.6 & 41.2 & 51.3 & 5.8 \\
Audio-Flamingo-3 & 30.3 & 43.0 & 53.4 & 4.3 \\
\midrule
Random & 25.0 & 39.3 & 50.4 & 4.2 \\
\bottomrule
\end{tabular}
\caption{VC Quality Ranking results.}
\label{tab:vc_ranking}
\vspace{-1em}
\end{table}

\subsection{LALMs as Voice Cloning Evaluators}
Beyond judging speaker consistency across dialogues, a natural downstream application is evaluating voice cloning systems: automatically ranking outputs from different models by their acoustic similarity to a reference speaker. We investigate whether LALMs can serve this role reliably, producing rankings that agree with human judgment.

To explore this, we construct a VC Quality Ranking task. From the benchmark dialogues, we randomly sample a pair of audio clips per dialogue: one serving as the reference speaker audio and the other as the source. We extract the transcript from the source clip and use it along with the reference audio to generate cloned outputs via three models — OpenVoice~\cite{qin2023openvoice}, YourTTS~\cite{casanova2022yourtts}, and CosyVoice3~\cite{du2025cosyvoice3} — yielding four candidates per sample (source, OpenVoice, YourTTS, CosyVoice3).
Three human annotators independently ranked 485 samples, with 50 shared samples for inter-annotator reliability (Kendall's W = 0.860). Human rankings serve as ground truth. Models are asked to rank all four candidates by acoustic similarity to the reference speaker. Evaluation protocol is the same as the Discrimination task (\S~\ref{evaluation_protocol}). The prompt is provided in Figure~\ref{fig:prompt_vc_ranking}.

As shown in Table~\ref{tab:vc_ranking}, Gemini-2.5-Pro achieves the strongest performance (Acc: 48.0\%, NDCG@2: 75.1\%), followed by Gemini-2.5-Flash and Qwen3-Omni-30B-A3B. However, most models perform modestly above the random baseline, and Exact Match remains low even for the best model (16.8\%). Unlike the Discrimination task, where candidates originate from distinct speakers, all candidates here are generated from the same source audio targeting the same speaker identity, requiring models to perceive much finer-grained acoustic variations. These results indicate that while some LALMs show promise as automatic voice cloning evaluators, reliable quality ranking aligned with human judgment remains a challenging open problem.

\section{Conclusion}
We present SpeakerSleuth, a benchmark for evaluating whether 
LALMs can reliably judge speaker consistency across multi-turn 
dialogues. Built around three complementary tasks that mirror 
real-world application requirements (Detection, Localization, 
and Discrimination), SpeakerSleuth comprises 1,818 
human-verified instances across four diverse datasets. Our 
evaluation reveals fundamental limitations: models lack stable 
internal thresholds, struggle with fine-grained turn-level 
analysis, and prioritize textual coherence over acoustic 
features. At the same time, they show a clear dissociation 
between detection and discrimination, indicating that acoustic 
discrimination capability is present but not effectively 
integrated into consistency judgment. These findings point to 
calibration, fine-grained reasoning, and modality integration 
as core challenges for building reliable audio-language judges.

\section*{Limitations}

SpeakerSleuth has several limitations. First, our benchmark covers only English dialogues. The synthetic generation pipeline itself is language-agnostic and 
can be extended to other languages. 
Second, our four datasets span diverse acoustic conditions 
(TV shows, meetings, synthesized speech, studio recordings), 
but we do not isolate individual acoustic factors such as 
background noise, reverberation, or recording quality. 
While our pipeline readily supports adding controlled 
perturbations via standard audio augmentation, systematic 
analysis of each factor is left to future work. Finally, we do not analyze how model performance varies across speaker demographics such as accents or age groups. 
Whether LALMs exhibit demographic biases in speaker 
consistency judgment remains an open question.
Despite these limitations, SpeakerSleuth provides a 
systematic framework for probing how LALMs reason about 
speaker identity in multi-turn dialogues, and the pipeline 
naturally accommodates future extensions along each of the 
axes above.

\section*{Acknowledgments\label{sec:acknowledgements}}
This work was supported by the National Research Foundation of Korea (NRF) grants (RS-2024-00333484 and RS-2024-00414981) and by the Institute of Information \& Communications Technology Planning \& Evaluation (IITP) grant (RS-2025-02215122, Development and Demonstration of Lightweight AI Model for Smart Homes), all funded by the Korean government (MSIT).

\bibliography{custom}

@inproceedings{Guo2023PromptttsCT,
  title={Prompttts: Controllable Text-To-Speech With Text Descriptions},
  author={Zhifang Guo and Yichong Leng and Yihan Wu and Sheng Zhao and Xu Tan},
  booktitle={{ICASSP 2023 - 2023 IEEE International Conference on Acoustics, Speech and Signal Processing (ICASSP)}},
  year={2023},
  pages={1-5},
  doi={10.1109/ICASSP49357.2023.10096285}
}

@inproceedings{xie-etal-2025-towards_speech_synthesis,
    title = "Towards Controllable Speech Synthesis in the Era of Large Language Models: A Systematic Survey",
    author = "Xie, Tianxin  and
      Rong, Yan  and
      Zhang, Pengfei  and
      Wang, Wenwu  and
      Liu, Li",
    editor = "Christodoulopoulos, Christos  and
      Chakraborty, Tanmoy  and
      Rose, Carolyn  and
      Peng, Violet",
    booktitle = "Proceedings of the 2025 Conference on Empirical Methods in Natural Language Processing",
    month = nov,
    year = "2025",
    address = "Suzhou, China",
    publisher = "Association for Computational Linguistics",
    url = "https://aclanthology.org/2025.emnlp-main.40/",
    doi = "10.18653/v1/2025.emnlp-main.40",
    pages = "764--791",
    ISBN = "979-8-89176-332-6"
}

@article{wang2023valle,
  title={Neural codec language models are zero-shot text to speech synthesizers},
  author={Wang, Chengyi and Chen, Sanyuan and Wu, Yu and Zhang, Ziqiang and Zhou, Long and Liu, Shujie and Chen, Zhuo and Liu, Yanqing and Wang, Huaming and Li, Jinyu and others},
  journal={arXiv preprint arXiv:2301.02111},
  year={2023}
}

@Inbook{speechevaluation,
author="Loizou, Philipos C.",
editor="Lin, Weisi
and Tao, Dacheng
and Kacprzyk, Janusz
and Li, Zhu
and Izquierdo, Ebroul
and Wang, Haohong",
title="Speech Quality Assessment",
bookTitle="Multimedia Analysis, Processing and Communications",
year="2011",
publisher="Springer Berlin Heidelberg",
address="Berlin, Heidelberg",
pages="623--654",
isbn="978-3-642-19551-8",
doi="10.1007/978-3-642-19551-8_23",
url="https://doi.org/10.1007/978-3-642-19551-8_23"
}

@inproceedings{
park2025longform,
title={Long-Form Speech Generation with Spoken Language Models},
author={Se Jin Park and Julian Salazar and Aren Jansen and Keisuke Kinoshita and Yong Man Ro and RJ Skerry-Ryan},
booktitle={Forty-second International Conference on Machine Learning},
year={2025},
url={https://openreview.net/forum?id=4AmFA0qNQ2}
}

@article{Mullennix1990Stimulus,
  title={Stimulus variability and processing dependencies in speech perception},
  author={John W. Mullennix and David B. Pisoni},
  journal={Perception \& Psychophysics},
  year={1990},
  volume={47},
  pages={379-390},
  url={https://api.semanticscholar.org/CorpusID:29489002}
}

@inproceedings{ju2024naturalspeech,
  title={NaturalSpeech 3: Zero-Shot Speech Synthesis with Factorized Codec and Diffusion Models},
  author={Ju, Zeqian and Wang, Yuancheng and Shen, Kai and Tan, Xu and Xin, Detai and Yang, Dongchao and Liu, Eric and Leng, Yichong and Song, Kaitao and Tang, Siliang and others},
  booktitle={International Conference on Machine Learning},
  pages={22605--22623},
  year={2024},
  organization={PMLR}
}

@article{li2025styletts,
  title={Styletts: A style-based generative model for natural and diverse text-to-speech synthesis},
  author={Li, Yinghao Aaron and Han, Cong and Mesgarani, Nima},
  journal={IEEE Journal of Selected Topics in Signal Processing},
  year={2025},
  publisher={IEEE}
}

@inproceedings{lee-etal-2025-behavior,
    title = "Behavior-{SD}: Behaviorally Aware Spoken Dialogue Generation with Large Language Models",
    author = "Lee, Sehun  and
      Kim, Kang-wook  and
      Kim, Gunhee",
    editor = "Chiruzzo, Luis  and
      Ritter, Alan  and
      Wang, Lu",
    booktitle = "Proceedings of the 2025 Conference of the Nations of the Americas Chapter of the Association for Computational Linguistics: Human Language Technologies (Volume 1: Long Papers)",
    month = apr,
    year = "2025",
    address = "Albuquerque, New Mexico",
    publisher = "Association for Computational Linguistics",
    url = "https://aclanthology.org/2025.naacl-long.484/",
    doi = "10.18653/v1/2025.naacl-long.484",
    pages = "9574--9593",
    ISBN = "979-8-89176-189-6"
}

@inproceedings{wang2017tacotron,
  title={Tacotron: Towards End-to-End Speech Synthesis},
  author={Wang, Yuxuan and Skerry-Ryan, RJ and Stanton, Daisy and Wu, Yonghui and Weiss, Ron J and Jaitly, Navdeep and Yang, Zongheng and Xiao, Ying and Chen, Zhifeng and Bengio, Samy and others},
  booktitle={Proc. Interspeech 2017},
  pages={4006--4010},
  year={2017}
}

@article{zhang2024covomix,
  title={CoVoMix: Advancing zero-shot speech generation for human-like multi-talker conversations},
  author={Zhang, Leying and Qian, Yao and Zhou, Long and Liu, Shujie and Wang, Dongmei and Wang, Xiaofei and Yousefi, Midia and Qian, Yanmin and Li, Jinyu and He, Lei and others},
  journal={Advances in Neural Information Processing Systems},
  volume={37},
  pages={100291--100317},
  year={2024}
}

@INPROCEEDINGS{melcepstral,
  author={Kubichek, R.},
  booktitle={Proceedings of IEEE Pacific Rim Conference on Communications Computers and Signal Processing}, 
  title={Mel-cepstral distance measure for objective speech quality assessment}, 
  year={1993},
  volume={1},
  number={},
  pages={125-128 vol.1},
  doi={10.1109/PACRIM.1993.407206}}

@inproceedings{saeki2022utmos,
  title={UTMOS: UTokyo-SaruLab System for VoiceMOS Challenge 2022},
  author={Saeki, Takaaki and Xin, Detai and Nakata, Wataru and Koriyama, Tomoki and Takamichi, Shinnosuke and Saruwatari, Hiroshi},
  booktitle={Interspeech 2022},
  year={2022},
  publisher={ISCA}
}

@inproceedings{chenaudio_speechjudge,
  title={Audio Large Language Models Can Be Descriptive Speech Quality Evaluators},
  author={Chen, Chen and Hu, Yuchen and Wang, Siyin and Wang, Helin and Chen, Zhehuai and Zhang, Chao and Yang, Chao-Han Huck and Chng, EngSiong},
  year = "2025",
  booktitle={The Thirteenth International Conference on Learning Representations}
}

@inproceedings{wang-etal-2025-qualispeech,
    title = "{Q}uali{S}peech: A Speech Quality Assessment Dataset with Natural Language Reasoning and Descriptions",
    author = "Wang, Siyin  and
      Yu, Wenyi  and
      Chen, Xianzhao  and
      Tian, Xiaohai  and
      Zhang, Jun  and
      Lu, Lu  and
      Tsao, Yu  and
      Yamagishi, Junichi  and
      Wang, Yuxuan  and
      Zhang, Chao",
    editor = "Che, Wanxiang  and
      Nabende, Joyce  and
      Shutova, Ekaterina  and
      Pilehvar, Mohammad Taher",
    booktitle = "Proceedings of the 63rd Annual Meeting of the Association for Computational Linguistics (Volume 1: Long Papers)",
    month = jul,
    year = "2025",
    address = "Vienna, Austria",
    publisher = "Association for Computational Linguistics",
    url = "https://aclanthology.org/2025.acl-long.1150/",
    doi = "10.18653/v1/2025.acl-long.1150",
    pages = "23588--23609",
    ISBN = "979-8-89176-251-0"
}

@inproceedings{wang2025enabling_speechjudge,
  title={Enabling auditory large language models for automatic speech quality evaluation},
  author={Wang, Siyin and Yu, Wenyi and Yang, Yudong and Tang, Changli and Li, Yixuan and Zhuang, Jimin and Chen, Xianzhao and Tian, Xiaohai and Zhang, Jun and Sun, Guangzhi and others},
  booktitle={ICASSP 2025-2025 IEEE International Conference on Acoustics, Speech and Signal Processing (ICASSP)},
  pages={1--5},
  year={2025},
  organization={IEEE}
}

@misc{wang2025speechllmasjudgesgeneralinterpretablespeech,
      title={SpeechLLM-as-Judges: Towards General and Interpretable Speech Quality Evaluation}, 
      author={Hui Wang and Jinghua Zhao and Yifan Yang and Shujie Liu and Junyang Chen and Yanzhe Zhang and Shiwan Zhao and Jinyu Li and Jiaming Zhou and Haoqin Sun and Yan Lu and Yong Qin},
      year={2025},
      eprint={2510.14664},
      archivePrefix={arXiv},
      primaryClass={cs.SD},
      url={https://arxiv.org/abs/2510.14664}, 
}

@inproceedings{lerner-etal-2022-bazinga,
    title = "Bazinga! A Dataset for Multi-Party Dialogues Structuring",
    author = {Lerner, Paul  and
      Bergo{\"e}nd, Juliette  and
      Guinaudeau, Camille  and
      Bredin, Herv{\'e}  and
      Maurice, Benjamin  and
      Lefevre, Sharleyne  and
      Bouteiller, Martin  and
      Berhe, Aman  and
      Galmant, L{\'e}o  and
      Yin, Ruiqing  and
      Barras, Claude},
    editor = "Calzolari, Nicoletta  and
      B{\'e}chet, Fr{\'e}d{\'e}ric  and
      Blache, Philippe  and
      Choukri, Khalid  and
      Cieri, Christopher  and
      Declerck, Thierry  and
      Goggi, Sara  and
      Isahara, Hitoshi  and
      Maegaard, Bente  and
      Mariani, Joseph  and
      Mazo, H{\'e}l{\`e}ne  and
      Odijk, Jan  and
      Piperidis, Stelios",
    booktitle = "Proceedings of the Thirteenth Language Resources and Evaluation Conference",
    month = jun,
    year = "2022",
    address = "Marseille, France",
    publisher = "European Language Resources Association",
    url = "https://aclanthology.org/2022.lrec-1.367/",
    pages = "3434--3441"
}

@article{du2024cosyvoice,
  title={Cosyvoice: A scalable multilingual zero-shot text-to-speech synthesizer based on supervised semantic tokens},
  author={Du, Zhihao and Chen, Qian and Zhang, Shiliang and Hu, Kai and Lu, Heng and Yang, Yexin and Hu, Hangrui and Zheng, Siqi and Gu, Yue and Ma, Ziyang and others},
  journal={arXiv preprint arXiv:2407.05407},
  year={2024}
}

@inproceedings{lee2023dailytalk,
  title={Dailytalk: Spoken dialogue dataset for conversational text-to-speech},
  author={Lee, Keon and Park, Kyumin and Kim, Daeyoung},
  booktitle={ICASSP 2023-2023 IEEE International Conference on Acoustics, Speech and Signal Processing (ICASSP)},
  pages={1--5},
  year={2023},
  organization={IEEE}
}

@inproceedings{carletta2005ami,
  title={The AMI meeting corpus: A pre-announcement},
  author={Carletta, Jean and Ashby, Simone and Bourban, Sebastien and Flynn, Mike and Guillemot, Mael and Hain, Thomas and Kadlec, Jaroslav and Karaiskos, Vasilis and Kraaij, Wessel and Kronenthal, Melissa and others},
  booktitle={International workshop on machine learning for multimodal interaction},
  pages={28--39},
  year={2005},
  organization={Springer}
}

@inproceedings{ecapa-tdnn,
  title     = {{ECAPA-TDNN}: Emphasized Channel Attention, Propagation and Aggregation in TDNN Based Speaker Verification},
  url       = {http://dx.doi.org/10.21437/Interspeech.2020-2650},
  doi       = {10.21437/interspeech.2020-2650},
  booktitle = {Interspeech 2020},
  publisher = {ISCA},
  author    = {Desplanques, Brecht and Thienpondt, Jenthe and Demuynck, Kris},
  year      = {2020},
  month     = oct,
  collection= {interspeech_2020}
}

@inproceedings{zhang2024comprehensive,
  title={A comprehensive analysis of the effectiveness of large language models as automatic dialogue evaluators},
  author={Zhang, Chen and D'Haro, Luis Fernando and Chen, Yiming and Zhang, Malu and Li, Haizhou},
  booktitle={Proceedings of the AAAI Conference on Artificial Intelligence},
  volume={38},
  number={17},
  pages={19515--19524},
  year={2024}
}

@inproceedings{li2023freevc,
  title={Freevc: Towards high-quality text-free one-shot voice conversion},
  author={Li, Jingyi and Tu, Weiping and Xiao, Li},
  booktitle={ICASSP 2023-2023 IEEE International Conference on Acoustics, Speech and Signal Processing (ICASSP)},
  pages={1--5},
  year={2023},
  organization={IEEE}
}

@article{furui1996overview,
  title={An overview of speaker recognition technology},
  author={Furui, Sadaoki},
  journal={Automatic Speech and Speaker Recognition: Advanced Topics},
  pages={31--56},
  year={1996},
  publisher={Springer}
}

@article{gish1994text,
  title={Text-independent speaker identification},
  author={Herbert Gish and Michael Schmidt},
  journal={IEEE signal processing magazine},
  volume={11},
  number={4},
  pages={18--32},
  year={1994},
  publisher={IEEE}
}

@misc{xu2025qwen25omnitechnicalreport,
      title={Qwen2.5-Omni Technical Report}, 
      author={Jin Xu and Zhifang Guo and Jinzheng He and Hangrui Hu and Ting He and Shuai Bai and Keqin Chen and Jialin Wang and Yang Fan and Kai Dang and Bin Zhang and Xiong Wang and Yunfei Chu and Junyang Lin},
      year={2025},
      eprint={2503.20215},
      archivePrefix={arXiv},
      primaryClass={cs.CL},
      url={https://arxiv.org/abs/2503.20215}, 
}

@misc{xu2025qwen3omnitechnicalreport,
      title={Qwen3-Omni Technical Report}, 
      author={Jin Xu and Zhifang Guo and Hangrui Hu and Yunfei Chu and Xiong Wang and Jinzheng He and Yuxuan Wang and Xian Shi and Ting He and Xinfa Zhu and Yuanjun Lv and Yongqi Wang and Dake Guo and He Wang and Linhan Ma and Pei Zhang and Xinyu Zhang and Hongkun Hao and Zishan Guo and Baosong Yang and Bin Zhang and Ziyang Ma and Xipin Wei and Shuai Bai and Keqin Chen and Xuejing Liu and Peng Wang and Mingkun Yang and Dayiheng Liu and Xingzhang Ren and Bo Zheng and Rui Men and Fan Zhou and Bowen Yu and Jianxin Yang and Le Yu and Jingren Zhou and Junyang Lin},
      year={2025},
      eprint={2509.17765},
      archivePrefix={arXiv},
      primaryClass={cs.CL},
      url={https://arxiv.org/abs/2509.17765}, 
}

@article{ye2025omnivinci,
  title={OmniVinci: Enhancing Architecture and Data for Omni-Modal Understanding LLM},
  author={Ye, Hanrong and Yang, Chao-Han Huck and Goel, Arushi and Huang, Wei and Zhu, Ligeng and Su, Yuanhang and Lin, Sean and Cheng, An-Chieh and Wan, Zhen and Tian, Jinchuan and others},
  journal={arXiv preprint arXiv:2510.15870},
  year={2025}
}

@misc{gemmateam2025gemma3technicalreport,
      title={Gemma 3 Technical Report}, 
      author={Gemma Team and Aishwarya Kamath and Johan Ferret and Shreya Pathak and Nino Vieillard and Ramona Merhej and Sarah Perrin and Tatiana Matejovicova and Alexandre Ramé and Morgane Rivière and Louis Rouillard and Thomas Mesnard and Geoffrey Cideron and Jean-bastien Grill and Sabela Ramos and Edouard Yvinec and Michelle Casbon and Etienne Pot and Ivo Penchev and Gaël Liu and Francesco Visin and Kathleen Kenealy and Lucas Beyer and Xiaohai Zhai and Anton Tsitsulin and Robert Busa-Fekete and Alex Feng and Noveen Sachdeva and Benjamin Coleman and Yi Gao and Basil Mustafa and Iain Barr and Emilio Parisotto and David Tian and Matan Eyal and Colin Cherry and Jan-Thorsten Peter and Danila Sinopalnikov and Surya Bhupatiraju and Rishabh Agarwal and Mehran Kazemi and Dan Malkin and Ravin Kumar and David Vilar and Idan Brusilovsky and Jiaming Luo and Andreas Steiner and Abe Friesen and Abhanshu Sharma and Abheesht Sharma and Adi Mayrav Gilady and Adrian Goedeckemeyer and Alaa Saade and Alex Feng and Alexander Kolesnikov and Alexei Bendebury and Alvin Abdagic and Amit Vadi and András György and André Susano Pinto and Anil Das and Ankur Bapna and Antoine Miech and Antoine Yang and Antonia Paterson and Ashish Shenoy and Ayan Chakrabarti and Bilal Piot and Bo Wu and Bobak Shahriari and Bryce Petrini and Charlie Chen and Charline Le Lan and Christopher A. Choquette-Choo and CJ Carey and Cormac Brick and Daniel Deutsch and Danielle Eisenbud and Dee Cattle and Derek Cheng and Dimitris Paparas and Divyashree Shivakumar Sreepathihalli and Doug Reid and Dustin Tran and Dustin Zelle and Eric Noland and Erwin Huizenga and Eugene Kharitonov and Frederick Liu and Gagik Amirkhanyan and Glenn Cameron and Hadi Hashemi and Hanna Klimczak-Plucińska and Harman Singh and Harsh Mehta and Harshal Tushar Lehri and Hussein Hazimeh and Ian Ballantyne and Idan Szpektor and Ivan Nardini and Jean Pouget-Abadie and Jetha Chan and Joe Stanton and John Wieting and Jonathan Lai and Jordi Orbay and Joseph Fernandez and Josh Newlan and Ju-yeong Ji and Jyotinder Singh and Kat Black and Kathy Yu and Kevin Hui and Kiran Vodrahalli and Klaus Greff and Linhai Qiu and Marcella Valentine and Marina Coelho and Marvin Ritter and Matt Hoffman and Matthew Watson and Mayank Chaturvedi and Michael Moynihan and Min Ma and Nabila Babar and Natasha Noy and Nathan Byrd and Nick Roy and Nikola Momchev and Nilay Chauhan and Noveen Sachdeva and Oskar Bunyan and Pankil Botarda and Paul Caron and Paul Kishan Rubenstein and Phil Culliton and Philipp Schmid and Pier Giuseppe Sessa and Pingmei Xu and Piotr Stanczyk and Pouya Tafti and Rakesh Shivanna and Renjie Wu and Renke Pan and Reza Rokni and Rob Willoughby and Rohith Vallu and Ryan Mullins and Sammy Jerome and Sara Smoot and Sertan Girgin and Shariq Iqbal and Shashir Reddy and Shruti Sheth and Siim Põder and Sijal Bhatnagar and Sindhu Raghuram Panyam and Sivan Eiger and Susan Zhang and Tianqi Liu and Trevor Yacovone and Tyler Liechty and Uday Kalra and Utku Evci and Vedant Misra and Vincent Roseberry and Vlad Feinberg and Vlad Kolesnikov and Woohyun Han and Woosuk Kwon and Xi Chen and Yinlam Chow and Yuvein Zhu and Zichuan Wei and Zoltan Egyed and Victor Cotruta and Minh Giang and Phoebe Kirk and Anand Rao and Kat Black and Nabila Babar and Jessica Lo and Erica Moreira and Luiz Gustavo Martins and Omar Sanseviero and Lucas Gonzalez and Zach Gleicher and Tris Warkentin and Vahab Mirrokni and Evan Senter and Eli Collins and Joelle Barral and Zoubin Ghahramani and Raia Hadsell and Yossi Matias and D. Sculley and Slav Petrov and Noah Fiedel and Noam Shazeer and Oriol Vinyals and Jeff Dean and Demis Hassabis and Koray Kavukcuoglu and Clement Farabet and Elena Buchatskaya and Jean-Baptiste Alayrac and Rohan Anil and Dmitry and Lepikhin and Sebastian Borgeaud and Olivier Bachem and Armand Joulin and Alek Andreev and Cassidy Hardin and Robert Dadashi and Léonard Hussenot},
      year={2025},
      eprint={2503.19786},
      archivePrefix={arXiv},
      primaryClass={cs.CL},
      url={https://arxiv.org/abs/2503.19786}, 
}

@misc{openai_gpt4o_2024,
  title        = {Hello GPT-4o},
  author       = {{OpenAI}},
  year         = {2024},
  howpublished = {\url{https://openai.com/index/hello-gpt-4o/}},
  note         = {Accessed 2026-04-19}
}

@misc{notebooklm,
  title        = {NotebookLM now lets you listen to a conversation about your sources},
  author       = {{Google}},
  year         = {2024},
  howpublished = {\url{https://blog.google/technology/ai/notebooklm-audio-overviews/}},
  note         = {Accessed 2026-04-19}
}

@misc{apple_intelligence,
  title        = {Apple Intelligence: AI for the rest of us.},
  author       = {{Apple}},
  year         = {2024},
  howpublished = {\url{https://www.apple.com/apple-intelligence/}},
  note         = {Accessed 2026-04-19}
}

@misc{microsoft2025phi4minitechnicalreportcompact,
      title={Phi-4-Mini Technical Report: Compact yet Powerful Multimodal Language Models via Mixture-of-LoRAs}, 
      author={Microsoft and : and Abdelrahman Abouelenin and Atabak Ashfaq and Adam Atkinson and Hany Awadalla and Nguyen Bach and Jianmin Bao and Alon Benhaim and Martin Cai and Vishrav Chaudhary and Congcong Chen and Dong Chen and Dongdong Chen and Junkun Chen and Weizhu Chen and Yen-Chun Chen and Yi-ling Chen and Qi Dai and Xiyang Dai and Ruchao Fan and Mei Gao and Min Gao and Amit Garg and Abhishek Goswami and Junheng Hao and Amr Hendy and Yuxuan Hu and Xin Jin and Mahmoud Khademi and Dongwoo Kim and Young Jin Kim and Gina Lee and Jinyu Li and Yunsheng Li and Chen Liang and Xihui Lin and Zeqi Lin and Mengchen Liu and Yang Liu and Gilsinia Lopez and Chong Luo and Piyush Madan and Vadim Mazalov and Arindam Mitra and Ali Mousavi and Anh Nguyen and Jing Pan and Daniel Perez-Becker and Jacob Platin and Thomas Portet and Kai Qiu and Bo Ren and Liliang Ren and Sambuddha Roy and Ning Shang and Yelong Shen and Saksham Singhal and Subhojit Som and Xia Song and Tetyana Sych and Praneetha Vaddamanu and Shuohang Wang and Yiming Wang and Zhenghao Wang and Haibin Wu and Haoran Xu and Weijian Xu and Yifan Yang and Ziyi Yang and Donghan Yu and Ishmam Zabir and Jianwen Zhang and Li Lyna Zhang and Yunan Zhang and Xiren Zhou},
      year={2025},
      eprint={2503.01743},
      archivePrefix={arXiv},
      primaryClass={cs.CL},
      url={https://arxiv.org/abs/2503.01743}, 
}

@misc{minicpm-o-2.6,
  title        = {MiniCPM-o 2.6: A GPT-4o Level MLLM for Vision, Speech, and Multimodal Live Streaming on Your Phone},
  author       = {{OpenBMB}},
  year         = {2025},
  howpublished = {\url{https://openbmb.vercel.app/minicpm-o-2-6-en}},
  note         = {Accessed 2026-04-19}
}

@article{defossez2024moshi,
  title={Moshi: a speech-text foundation model for real-time dialogue},
  author={D{\'e}fossez, Alexandre and Mazar{\'e}, Laurent and Orsini, Manu and Royer, Am{\'e}lie and P{\'e}rez, Patrick and J{\'e}gou, Herv{\'e} and Grave, Edouard and Zeghidour, Neil},
  journal={arXiv preprint arXiv:2410.00037},
  year={2024}
}

@inproceedings{
zhang2025covomix2,
title={CoVoMix2: Advancing Zero-Shot Dialogue Generation with Fully Non-Autoregressive Flow Matching},
author={Leying Zhang and Yao Qian and Xiaofei Wang and Manthan Thakker and Dongmei Wang and Jianwei Yu and Haibin Wu and Yuxuan Hu and Jinyu Li and Yanmin Qian and sheng zhao},
booktitle={The Thirty-ninth Annual Conference on Neural Information Processing Systems},
year={2025},
url={https://openreview.net/forum?id=0fq8vYOnxi}
}

@misc{yang2025qwen3technicalreport,
      title={Qwen3 Technical Report}, 
      author={An Yang and Anfeng Li and Baosong Yang and Beichen Zhang and Binyuan Hui and Bo Zheng and Bowen Yu and Chang Gao and Chengen Huang and Chenxu Lv and Chujie Zheng and Dayiheng Liu and Fan Zhou and Fei Huang and Feng Hu and Hao Ge and Haoran Wei and Huan Lin and Jialong Tang and Jian Yang and Jianhong Tu and Jianwei Zhang and Jianxin Yang and Jiaxi Yang and Jing Zhou and Jingren Zhou and Junyang Lin and Kai Dang and Keqin Bao and Kexin Yang and Le Yu and Lianghao Deng and Mei Li and Mingfeng Xue and Mingze Li and Pei Zhang and Peng Wang and Qin Zhu and Rui Men and Ruize Gao and Shixuan Liu and Shuang Luo and Tianhao Li and Tianyi Tang and Wenbiao Yin and Xingzhang Ren and Xinyu Wang and Xinyu Zhang and Xuancheng Ren and Yang Fan and Yang Su and Yichang Zhang and Yinger Zhang and Yu Wan and Yuqiong Liu and Zekun Wang and Zeyu Cui and Zhenru Zhang and Zhipeng Zhou and Zihan Qiu},
      year={2025},
      eprint={2505.09388},
      archivePrefix={arXiv},
      primaryClass={cs.CL},
      url={https://arxiv.org/abs/2505.09388}, 
}

@inproceedings{
ju2025mooncast,
title={MoonCast: High-Quality Zero-Shot Podcast Generation},
author={Zeqian Ju and Dongchao Yang and Kai Shen and Yichong Leng and Zhengtao Wang and Songxiang Liu and Xinyu Zhou and Tao Qin and Xiangyang Li and Jianwei Yu and Xu Tan},
booktitle={The Thirty-ninth Annual Conference on Neural Information Processing Systems},
year={2025},
url={https://openreview.net/forum?id=MVlKSYR7HX}
}

@INPROCEEDINGS{pesq,
  author={Rix, A.W. and Beerends, J.G. and Hollier, M.P. and Hekstra, A.P.},
  booktitle={2001 IEEE International Conference on Acoustics, Speech, and Signal Processing. Proceedings (Cat. No.01CH37221)}, 
  title={Perceptual evaluation of speech quality (PESQ)-a new method for speech quality assessment of telephone networks and codecs}, 
  year={2001},
  volume={2},
  number={},
  pages={749-752 vol.2},
  doi={10.1109/ICASSP.2001.941023}}

@INPROCEEDINGS{human_mos,
  author={Ribeiro, Flávio and Florêncio, Dinei and Zhang, Cha and Seltzer, Michael},
  booktitle={2011 IEEE International Conference on Acoustics, Speech and Signal Processing (ICASSP)}, 
  title={CROWDMOS: An approach for crowdsourcing mean opinion score studies}, 
  year={2011},
  volume={},
  number={},
  pages={2416-2419},
  doi={10.1109/ICASSP.2011.5946971}}

@article{wang2025paymoreattentiontoaudio,
  title={Pay More Attention To Audio: Mitigating Imbalance of Cross-Modal Attention in Large Audio Language Models},
  author={Wang, Junyu and Ma, Ziyang and Luo, Zhengding and Wang, Tianrui and Ge, Meng and Wang, Xiaobao and Wang, Longbiao},
  journal={arXiv preprint arXiv:2509.18816},
  year={2025}
}

@misc{google_gemini25flash_2025,
  title        = {Continuing to bring you our latest models, with an improved Gemini 2.5 Flash and Flash-Lite release},
  author       = {Basu Mallick, Shrestha and Lall, Sid and Gleicher, Zach and Olszewska, Kate},
  year         = {2025},
  month        = {September},
  howpublished = {\url{https://developers.googleblog.com/en/continuing-to-bring-you-our-latest-models-with-an-improved-gemini-2-5-flash-and-flash-lite-release/}},
  note         = {Google Developers Blog; Accessed 2026-04-19}
}

@misc{google_gemini25pro_2025,
  title        = {Gemini 2.5: Our most intelligent AI model},
  author       = {Koray Kavukcuoglu},
  year         = {2025},
  month        = {March},
  howpublished = {\url{https://blog.google/innovation-and-ai/models-and-research/google-deepmind/gemini-model-thinking-updates-march-2025/}},
  note         = {Google Developers Blog; Accessed 2026-04-19}
}

@article{mcinnes2018umap,
  title={{UMAP}: Uniform Manifold Approximation and Projection},
  author={McInnes, Leland and Healy, John and Saul, Nathaniel and Gro{\ss}berger, Lukas},
  journal={Journal of Open Source Software},
  volume={3},
  number={29},
  year={2018}
}

@article{chen2022wavlm,
  title={Wavlm: Large-scale self-supervised pre-training for full stack speech processing},
  author={Chen, Sanyuan and Wang, Chengyi and Chen, Zhengyang and Wu, Yu and Liu, Shujie and Chen, Zhuo and Li, Jinyu and Kanda, Naoyuki and Yoshioka, Takuya and Xiao, Xiong and others},
  journal={IEEE Journal of Selected Topics in Signal Processing},
  volume={16},
  number={6},
  pages={1505--1518},
  year={2022},
  publisher={IEEE}
}

@Article{clustering,
AUTHOR = {Khoma, Volodymyr and Khoma, Yuriy and Brydinskyi, Vitalii and Konovalov, Alexander},
TITLE = {Development of Supervised Speaker Diarization System Based on the PyAnnote Audio Processing Library},
JOURNAL = {Sensors},
VOLUME = {23},
YEAR = {2023},
NUMBER = {4},
ARTICLE-NUMBER = {2082},
URL = {https://www.mdpi.com/1424-8220/23/4/2082},
PubMedID = {36850680},
ISSN = {1424-8220},
DOI = {10.3390/s23042082}
}

@inproceedings{all-pair,
  title     = {Deep Neural Network Embeddings for Text-Independent Speaker Verification},
  author    = {David Snyder and Daniel Garcia-Romero and Daniel Povey and Sanjeev Khudanpur},
  year      = {2017},
  booktitle = {Interspeech 2017},
  pages     = {999--1003},
  doi       = {10.21437/Interspeech.2017-620},
  issn      = {2958-1796},
}

@inproceedings{pezeshkpour2024mcq_positional_bias,
  title={Large language models sensitivity to the order of options in multiple-choice questions},
  author={Pezeshkpour, Pouya and Hruschka, Estevam},
  booktitle={Findings of the Association for Computational Linguistics: NAACL 2024},
  pages={2006--2017},
  year={2024}
}

@article{borsos2023audiolm,
  title={Audiolm: a language modeling approach to audio generation},
  author={Borsos, Zal{\'a}n and Marinier, Rapha{\"e}l and Vincent, Damien and Kharitonov, Eugene and Pietquin, Olivier and Sharifi, Matt and Roblek, Dominik and Teboul, Olivier and Grangier, David and Tagliasacchi, Marco and others},
  journal={IEEE/ACM transactions on audio, speech, and language processing},
  volume={31},
  pages={2523--2533},
  year={2023},
  publisher={IEEE}
}

@inproceedings{reimers-gurevych-2019-sentence,
    title = "Sentence-{BERT}: Sentence Embeddings using {S}iamese {BERT}-Networks",
    author = "Reimers, Nils  and
      Gurevych, Iryna",
    editor = "Inui, Kentaro  and
      Jiang, Jing  and
      Ng, Vincent  and
      Wan, Xiaojun",
    booktitle = "Proceedings of the 2019 Conference on Empirical Methods in Natural Language Processing and the 9th International Joint Conference on Natural Language Processing (EMNLP-IJCNLP)",
    month = nov,
    year = "2019",
    address = "Hong Kong, China",
    publisher = "Association for Computational Linguistics",
    url = "https://aclanthology.org/D19-1410/",
    doi = "10.18653/v1/D19-1410",
    pages = "3982--3992"
}

@misc{WatchTheSkies2025,
  title        = {Watch the Skies},
  author       = {Victor Danell},
  year         = {2025},
  howpublished = {\url{https://www.imdb.com/title/tt14807348/}},
  note         = {Accessed 2026-04-19}
}

@article{lee2026spokenus,
  title={SpokenUS: A Spoken User Simulator for Task-Oriented Dialogue},
  author={Lee, Jonggeun and Pyo, Junseong and Park, Jeongmin and Jo, Yohan},
  journal={arXiv preprint arXiv:2603.16783},
  year={2026}
}

@article{qin2023openvoice,
  title={Openvoice: Versatile instant voice cloning},
  author={Qin, Zengyi and Zhao, Wenliang and Yu, Xumin and Sun, Xin},
  journal={arXiv preprint arXiv:2312.01479},
  year={2023}
}

@inproceedings{casanova2022yourtts,
  title={Yourtts: Towards zero-shot multi-speaker tts and zero-shot voice conversion for everyone},
  author={Casanova, Edresson and Weber, Julian and Shulby, Christopher D and Junior, Arnaldo Candido and G{\"o}lge, Eren and Ponti, Moacir A},
  booktitle={International conference on machine learning},
  pages={2709--2720},
  year={2022},
  organization={PMLR}
}

@inproceedings{
ghosh2025audioflamingo3,
title={Audio Flamingo 3: Advancing Audio Intelligence with Fully Open Large Audio Language Models},
author={Sreyan Ghosh and Arushi Goel and Jaehyeon Kim and Sonal Kumar and Zhifeng Kong and Sang-gil Lee and Chao-Han Huck Yang and Ramani Duraiswami and Dinesh Manocha and Rafael Valle and Bryan Catanzaro},
booktitle={The Thirty-ninth Annual Conference on Neural Information Processing Systems},
year={2025},
url={https://openreview.net/forum?id=FjByDpDVIO}
}

@article{du2025cosyvoice3,
  title={Cosyvoice 3: Towards in-the-wild speech generation via scaling-up and post-training},
  author={Du, Zhihao and Gao, Changfeng and Wang, Yuxuan and Yu, Fan and Zhao, Tianyu and Wang, Hao and Lv, Xiang and Wang, Hui and Ni, Chongjia and Shi, Xian and others},
  journal={arXiv preprint arXiv:2505.17589},
  year={2025}
}
\newpage
\appendix
\section{SpeakerSleuth Details}
\label{app:speakersleuth_details}

\begin{table}[h]
\centering
\caption{Breakdown of TV series and episodes used (all from Season 1) from Bazinga dataset in SpeakerSleuth.}
\label{tab:bazinga_details}
\small
\begin{tabular}{lc}
\toprule
\textbf{TV Series/Movie} & \textbf{Episodes} \\
\midrule
24 & 24 \\
Battlestar Galactica & 13 \\
Breaking Bad & 7 \\
Buffy The Vampire Slayer & 12 \\
ER & 25 \\
Friends & 24 \\
Game of Thrones & 10 \\
Homeland & 12 \\
Lost & 25 \\
Six Feet Under & 13 \\
Star Wars & 7 \\
The Big Bang Theory & 17 \\
The Office & 6 \\
The Walking Dead & 6 \\
\midrule
\textbf{Total} & \textbf{201 episodes} \\
\bottomrule
\end{tabular}
\end{table}
\subsection{Dataset Details}
\label{app:dataset_details}

We use four datasets for SpeakerSleuth. This section describes 
the specific subset used from each and its licensing. 
Table~\ref{tab:benchmark_stats} summarizes the resulting 
statistics.

\paragraph{Bazinga.}
Bazinga~\cite{lerner-etal-2022-bazinga} is a multi-party dialogue 
dataset from TV series and movies. We use 14 series spanning 201 
episodes total, covering comedy, drama, and documentary formats 
to capture diverse speaking styles. 
Table~\ref{tab:bazinga_details} provides the per-series breakdown.

\paragraph{AMI Meeting Corpus.}
The AMI Meeting Corpus~\cite{carletta2005ami} consists of 100 
hours of meeting recordings across three rooms, predominantly 
featuring non-native speakers. We use the evaluation set of 16 
meetings, which provides challenging real-world acoustic 
conditions: spontaneous disfluencies, overlapping speech, 
variable room acoustics, and non-native accents.

\paragraph{Behavior-SD.}
Behavior-SD~\cite{lee-etal-2025-behavior} is a large-scale dataset 
of synthesized dialogues (100K+ dialogues, 2,164 hours) with 
annotations for conversational behaviors such as fillers, 
backchannels, and interruptions. We use the test set of 925 
dialogues. Behavior-SD serves as a control condition with clean, 
synthesized speech against the more challenging real-world 
recordings from other datasets.

\paragraph{DailyTalk.}
DailyTalk~\cite{lee2023dailytalk} is a high-quality conversational 
TTS dataset derived from DailyDialog, containing 2,541 studio-quality 
dialogues between two participants (one male, one female). We use 
all 2,541 dialogues. While the speaker diversity is limited, the 
consistent studio conditions enable evaluation of within-speaker 
consistency across varied conversational scenarios.

\paragraph{Licensing.}
All datasets are used under their respective licenses: AMI Corpus 
and Behavior-SD under CC BY 4.0, Bazinga under CC BY-NC 4.0, and 
DailyTalk under CC BY-SA 4.0. Our use for benchmark evaluation is 
consistent with their intended academic research purposes.

\subsection{Verification Details}
\label{app:verification_details}

This section details the verification pipeline outlined in 
Section~\ref{verification}. For text-based filtering, we use 
Qwen3-32B~\cite{yang2025qwen3technicalreport} with the prompt 
shown in Figure~\ref{fig:prompt_coherence_filter}. For 
audio-based verification, three annotators from our research 
team, all with expertise in speech processing and audio 
evaluation, evaluated all samples based on the criteria below. 
Annotators also confirmed that each reference audio shares 
consistent speaker identity and acoustic environment with the 
target speaker's turns in the dialogue, ensuring it serves as 
a reliable anchor for acoustic comparison.

\paragraph{Audio Quality.} 
Annotators checked for excessive background noise interfering 
with speaker characteristics, confirmed clear human speech in 
all utterances, and flagged excessive clipping or silence.

\paragraph{Naturalness.} 
Voice-converted turns were checked for robotic artifacts, and annotators verified that pitch, timbre, tone, and emotion flowed naturally throughout each turn.

\paragraph{Scenario Validity.} 
For S1, annotators confirmed consistent speaker identity and 
acoustic environment across all turns. For S2, they verified 
clear gender distinction between the original and converted 
speaker. For S3, they ensured the substituted speaker was 
acoustically distinguishable from the target, despite the 
intended similarity.

\subsection{Dialogue and Speaker Embedding Visualization Details}
\label{app:embedding_details}
The visualization in Figure~\ref{fig:speaker_embedding} was generated using the UMAP~\cite{mcinnes2018umap} algorithm with \texttt{n\_neighbors} set to 15. For dialogue embeddings, we employed the all-MiniLM-L6-v2~\cite{reimers-gurevych-2019-sentence} text embedding model, while speaker embeddings were extracted using ECAPA-TDNN~\cite{ecapa-tdnn}.

\section{Experimental Setup Details}
\label{app:experimental_setup}

\subsection{LALM Judges}

Table~\ref{tab:model_details} summarizes the twelve LALMs we 
evaluate, along with their parameter counts. All inference is 
conducted with temperature 0 on NVIDIA A100 80GB GPUs using 
CUDA 12.4.

\begin{table}[h]
\centering
\small
\begin{tabular}{ll}
\toprule
\textbf{Model} & \textbf{Parameters} \\
\midrule
\multicolumn{2}{l}{\textit{Proprietary}} \\
GPT-4o-audio          & N/A \\
Gemini-2.5-Pro        & N/A \\
Gemini-2.5-Flash      & N/A \\
Gemini-2.5-Flash-Lite & N/A \\
\midrule
\multicolumn{2}{l}{\textit{Open-source}} \\
Qwen2.5-Omni-3B       & 3B \\
Qwen2.5-Omni-7B       & 7B \\
Qwen3-Omni-30B-A3B    & 30B total, 3B active \\
MiniCPM-o-2.6         & 8B \\
Gemma-3n-E4B          & 8B total, 4B effective \\
Phi-4-multimodal      & 5.6B \\
OmniVinci             & 9B \\
Audio-Flamingo-3      & 7B \\
\bottomrule
\end{tabular}
\caption{LALM judges evaluated in our benchmark. ``N/A'' 
indicates parameter counts not publicly disclosed.}
\label{tab:model_details}
\end{table}

\subsection{Speaker Embedding Methods}
\label{embedding_methods_details}

We describe the three speaker embedding methods introduced in 
Section~\ref{sec:speaker_embedding}. For each target speaker audio set 
$\mathcal{A}_S = \{a_i\}_{i \in I}$, we first extract per-turn 
speaker embeddings $e_i$ from $a_i$ using either 
WavLM~\cite{chen2022wavlm} or ECAPA-TDNN~\cite{ecapa-tdnn}. 
The three methods differ in how they aggregate these embeddings 
to produce a consistency score per turn and a similarity score 
per candidate.

\paragraph{Pairwise Similarity.} 
The per-turn consistency score is the mean pairwise cosine 
similarity:
\begin{equation*}
s_i = \tfrac{1}{|I|-1}\sum_{j \neq i} \operatorname{sim}(e_i, e_j).
\end{equation*}
Turn $i$ is flagged if $s_i < \tau_{\text{pair}}$. For 
\textit{Discrimination}, each candidate $o_j$ is scored as
\begin{equation*}
q_j = \tfrac{1}{|I|}\sum_{i \in I} \operatorname{sim}(o_j, e_i).
\end{equation*}

\paragraph{Centroid Distance.} 
With context centroid $c = \tfrac{1}{|I|}\sum_{i \in I} e_i$ 
and $\operatorname{dist}(x,y) = 1 - \operatorname{sim}(x,y)$, 
the per-turn score is
\begin{equation*}
s_i = \operatorname{dist}(e_i, c).
\end{equation*}
Turn $i$ is flagged if $s_i > \tau_{\text{cent}}$. For 
\textit{Discrimination}, each candidate is scored as
\begin{equation*}
q_j = \operatorname{dist}(o_j, c).
\end{equation*}

\paragraph{Reference Comparison.} 
Given a reference embedding $r$ extracted from the target 
speaker's reference audio, the per-turn score is
\begin{equation*}
s_i = \operatorname{sim}(e_i, r).
\end{equation*}
Turn $i$ is flagged if $s_i < \tau_{\text{ref}}$. For 
\textit{Discrimination}, each candidate is scored as
\begin{equation*}
q_j = \operatorname{sim}(o_j, r).
\end{equation*}

\paragraph{Task Application.} 
Given these per-method scores, we apply the three tasks 
uniformly. \textit{Detection} classifies $\mathcal{A}_S$ as 
inconsistent if any turn is flagged. \textit{Localization} 
outputs the flagged turn(s). \textit{Discrimination} returns 
$\arg\max_j q_j$ (or $\arg\min$ for Centroid Distance) for 
the classification formulation, and $\{q_j\}$ sorted in 
descending order (ascending for Centroid) for the ranking 
formulation.

\begin{figure*}[ht]
\centering
\includegraphics[width=0.95\textwidth]{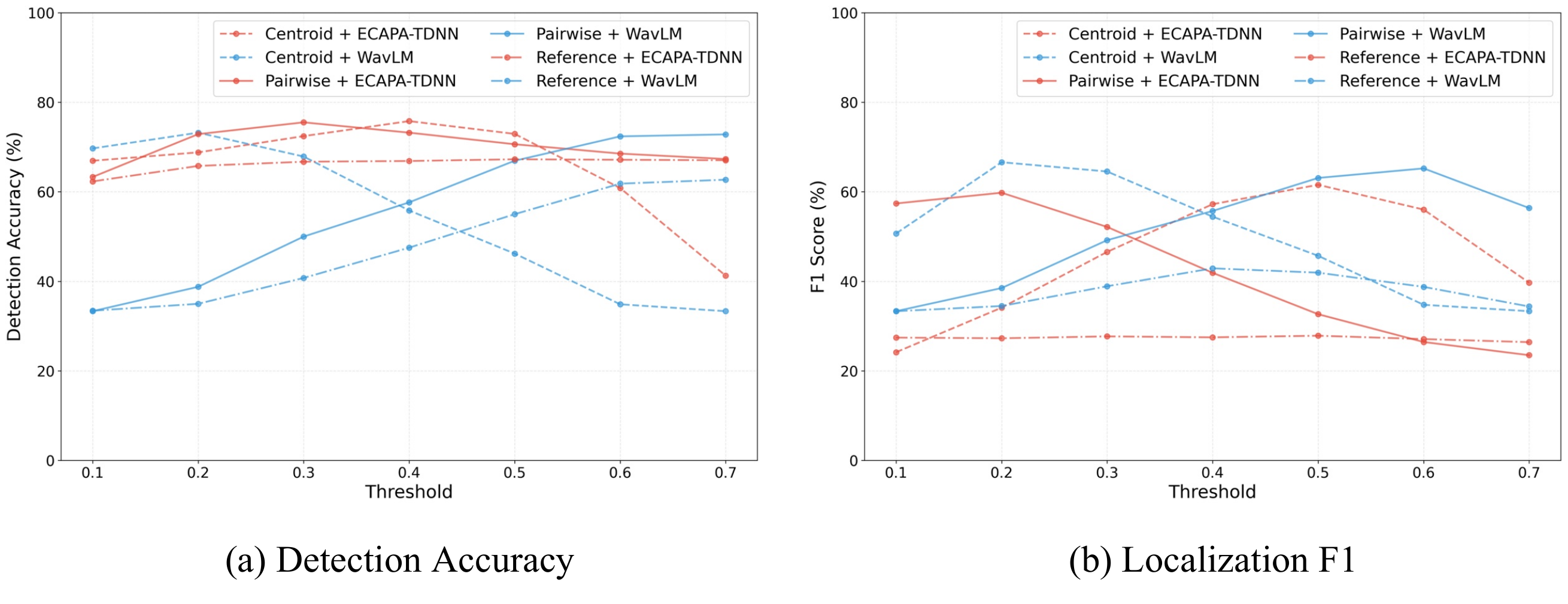}
\caption{Detection and Localization Performance with varying thresholds $\tau$.}
\label{fig:threshold_sensitivity}
\end{figure*}

\paragraph{Threshold Sensitivity Analysis}

We sweep $\tau$ from 0.1 to 0.7 in 0.1 increments for each of 
the six method-extractor combinations and evaluate detection 
accuracy and localization F1-score. As 
Figure~\ref{fig:threshold_sensitivity} shows, optimal 
thresholds vary substantially across configurations. Our 
chosen values ($\tau_{\text{pair}} = \tau_{\text{ref}} = 0.4$, 
$\tau_{\text{cent}} = 0.3$) provide reasonable performance 
across both backbones.

\subsection{Evaluation Metrics}
\label{evaluation_metrics}

This section defines the evaluation metrics used for the three 
tasks.

\paragraph{Task 1: Detection.} 
Detection is a binary classification problem where the model 
predicts whether a dialogue is consistent or inconsistent. 
Since Scenario 1 (S1) contains only fully consistent 
dialogues and Scenarios 2 and 3 (S2, S3) contain only 
dialogues with an inconsistent turn, per-scenario accuracy 
reduces to the proportion of dialogues correctly classified 
within each scenario:
\begin{equation*}
\text{Acc}_{\text{S}} = \frac{1}{N_{\text{S}}} \sum_{d \in \text{S}} \mathbb{1}[\hat{y}_d = y_d],
\end{equation*}
where $N_{\text{S}}$ is the number of dialogues in scenario 
$\text{S} \in \{\text{S1}, \text{S2}, \text{S3}\}$, 
$\hat{y}_d$ is the prediction, and $y_d$ is the ground-truth 
label. Reporting accuracy per scenario reveals threshold 
calibration patterns that would be hidden under a single 
aggregate score.

\paragraph{Task 2: Localization.} 
Given all target speaker turns at once, the model outputs the 
set of turns it judges to be inconsistent (or \texttt{None}). 
We evaluate this output as a multi-label classification: for 
each sample, let $\hat{G}$ be the predicted set and $G$ the 
ground-truth set of inconsistent turns. Per-sample precision 
and recall are 
$\text{P} = |\hat{G} \cap G| / |\hat{G}|$ and 
$\text{R} = |\hat{G} \cap G| / |G|$, with F1 defined as
\begin{equation*}
\text{F1} = \frac{2\,\text{P} \cdot \text{R}}{\text{P} + \text{R}}.
\end{equation*}
We use the convention $\text{P} = 0$ when $|\hat{G}| = 0$, 
$\text{R} = 0$ when $|G| = 0$, and $\text{F1} = 1$ when 
$|\hat{G}| = |G| = 0$. All metrics are macro-averaged across 
samples.

\paragraph{Task 3: Discrimination.} 
Discrimination presents three candidates, one from each 
scenario (S1, S2, S3), in a randomly shuffled order to avoid 
positional bias. We evaluate two formulations.

\textit{Classification.} The model selects the single 
best-matching candidate (i.e., S1, the original). We report 
accuracy over $N$ samples:
\begin{equation*}
\text{Acc} = \frac{1}{N}\sum_{j=1}^{N} \mathbb{1}[\hat{c}_j = c_j],
\end{equation*}
where $\hat{c}_j$ is the predicted candidate and $c_j$ is the 
ground-truth candidate (i.e., S1).

\textit{Ranking.} The model orders all three candidates by 
acoustic similarity to the target speaker. The ground-truth 
ordering is $\text{S1} \succ \text{S3} \succ \text{S2}$: S1 is 
the original consistent clip; S3 is voice-converted from the 
most acoustically similar speaker; and S2 is voice-converted 
from a speaker of a different gender, the most dissimilar 
option. Relevance scores are assigned as 2, 1, and 0 for ranks 
1, 2, and 3, respectively. Per sample, NDCG@$k$ is computed as
\begin{equation*}
\text{NDCG@}k = \frac{\text{DCG}@k}{\text{IDCG}@k},
\end{equation*}
\begin{equation*}
\text{DCG}@k = \sum_{i=1}^{k} \frac{2^{\,\text{rel}_i} - 1}{\log_2(i+1)}
\end{equation*}
where $\text{rel}_i$ is the relevance score of the candidate 
at rank $i$ and $\text{IDCG}@k$ is the DCG of the ideal 
ranking. NDCG@$k$ is averaged across samples. We additionally 
report Exact Match, the proportion of samples where the full 
predicted ranking matches the ground truth.

\begin{table*}[!ht]
\centering
\scriptsize
\setlength{\tabcolsep}{4pt}
\begin{tabular}{l|cccc|c|ccc|ccc|c|ccccc}
\toprule
& \multicolumn{4}{c|}{\textbf{Detection}} & \multicolumn{8}{c|}{\textbf{Localization}} & \multicolumn{5}{c}{\textbf{Discrimination}} \\
\cmidrule(lr){2-5} \cmidrule(lr){6-13} \cmidrule(lr){14-18}
\textbf{Model} & S1 & S2 & S3 & Bal & \multicolumn{1}{c|}{S1} & \multicolumn{3}{c|}{S2} & \multicolumn{3}{c|}{S3} & Bal & \multicolumn{5}{c}{Classification} \\
\cmidrule(lr){2-5} \cmidrule(lr){6-6} \cmidrule(lr){7-9} \cmidrule(lr){10-12} \cmidrule(lr){13-13} \cmidrule(lr){14-18}
& \multicolumn{4}{c|}{Acc} & F1 & P & R & F1 & P & R & F1 & F1 & Baz & AMI & B-SD & Daily & Avg \\
\midrule
\rowcolor{yellow!30}
\multicolumn{18}{c}{\textit{Per-Turn Pairwise}} \\
GPT-4o-audio       & 0.2  & 99.8  & 100.0 & 50.0 & 0.2  & 20.9 & 98.6 & 34.3 & 20.4 & 97.9 & 33.7 & 17.1 & 55.9 & 43.5 & 40.3 & 30.7 & 43.0 \\
Gemini-2.5-Pro      & 18.0 & 95.0  & 91.0  & \underline{55.5} & 18.0 & 38.2 & 82.4 & 48.8 & 27.8 & 67.5 & 36.7 & 30.4 & 41.7 & 69.6 & 54.4 & 51.9 & 50.3 \\
Gemini-2.5-Flash    & 11.6 & 96.4  & 92.9  & 53.1 & 11.6 & 38.6 & 89.6 & 50.4 & 28.3 & 72.2 & 38.2 & 28.0 & 47.6 & 80.4 & 76.1 & 74.1 & \textbf{65.8} \\
Gemini-2.5-Flash-Lite & 4.1 & 97.9 & 98.6  & 51.2 & 4.1  & 26.9 & 86.0 & 39.1 & 24.4 & 80.5 & 35.9 & 20.8 & 30.7 & 34.8 & 47.2 & 57.7 & 43.7 \\
Qwen2.5-Omni-3B    & 0.0  & 100.0 & 100.0 & 50.0 & 0.0  & 20.4 & 99.3 & 33.8 & 20.3 & 98.4 & 33.5 & 16.8 & 53.8 & 47.8 & 45.9 & 45.5 & 48.7 \\
Qwen2.5-Omni-7B    & 0.7  & 100.0 & 100.0 & 50.3 & 0.7  & 21.8 & 98.1 & 35.3 & 21.6 & 96.0 & 34.7 & 17.9 & 44.8 & 30.4 & 34.0 & 31.2 & 36.6 \\
Qwen3-Omni-30B-A3B & 2.9  & 99.5  & 98.6  & 51.0 & 2.9  & 25.7 & 96.0 & 39.1 & 22.0 & 87.6 & 34.3 & 19.8 & 79.2 & 76.1 & 46.5 & 44.4 & \underline{59.6} \\
MiniCPM-o-2.6      & 85.3 & 35.6 & 17.8 & \textbf{56.0} & 85.3 & 25.2 & 28.3 & 26.1 & 6.4  & 7.6  & 6.7  & \textbf{50.9} & 36.3 & 30.4 & 40.9 & 42.3 & 38.9 \\
Gemma-3n-E4B       & 33.9 & 68.2  & 67.2  & 50.8 & 33.9 & 13.3 & 30.9 & 17.4 & 13.6 & 31.1 & 17.7 & 25.7 & 33.0 & 34.8 & 39.6 & 42.3 & 37.8 \\
Phi-4-multimodal   & 11.4 & 90.2  & 87.9  & 50.2 & 11.4 & 21.9 & 59.1 & 30.3 & 19.6 & 54.9 & 27.4 & 20.1 & 31.6 & 30.4 & 37.1 & 41.3 & 36.0 \\
Omnivinci          & 0.3  & 100.0 & 99.5 & 50.0 & 0.3  & 23.2 & 99.3 & 37.1 & 22.0 & 94.5 & 35.1 & 18.2 & 41.5 & 34.8 & 38.4 & 41.3 & 40.1 \\
Audio-Flamingo-3   & 75.3 & 27.7  & 23.2  & 50.4 & 75.3 & 5.6  & 17.3 & 7.9  & 4.4  & 14.9 & 6.4  & \underline{41.2} & 33.0 & 30.4 & 36.5 & 41.3 & 36.3 \\
\midrule
\rowcolor{blue!20}
\multicolumn{18}{c}{\textit{Multi-Turn (from Table~\ref{tab:main_results})}} \\
GPT-4o-audio       & 72.9 & 32.8  & 29.5  & 52.0 & 71.5 & 12.6 & 25.9 & 15.0 & 8.7  & 19.2 & 11.1 & 42.3 & 50.2 & 30.4 & 38.4 & 34.4 & 40.7 \\
Gemini-2.5-Pro      & 73.9 & 71.6  & 39.3  & \textbf{64.7} & 65.2 & 59.3 & 70.3 & 62.5 & 44.0 & 56.6 & 47.5 & \textbf{60.1} & 77.8 & 76.1 & 81.1 & 87.2 & \textbf{81.5} \\
Gemini-2.5-Flash    & 97.4 & 45.5  & 12.0  & \underline{63.1} & 64.4 & 55.6 & 90.1 & 62.3 & 38.6 & 72.4 & 45.0 & \underline{59.0} & 70.8 & 73.9 & 69.2 & 86.8 & \underline{75.6} \\
Gemini-2.5-Flash-Lite & 40.8 & 70.3 & 69.3 & 55.3 & 0.3  & 25.8 & 94.7 & 36.4 & 24.5 & 94.1 & 35.4 & 18.1 & 47.6 & 45.7 & 47.8 & 44.4 & 46.5 \\
Qwen2.5-Omni-3B    & 54.5 & 49.3  & 47.8  & 51.5 & 21.8 & 45.6 & 91.6 & 47.7 & 28.9 & 81.6 & 36.0 & 31.8 & 42.9 & 39.1 & 41.5 & 38.1 & 40.8 \\
Qwen2.5-Omni-7B    & 32.0 & 72.3  & 69.8  & 51.5 & 38.0 & 14.1 & 61.8 & 22.6 & 13.2 & 57.5 & 21.0 & 29.9 & 57.5 & 30.4 & 39.6 & 31.7 & 42.7 \\
Qwen3-Omni-30B-A3B & 88.1 & 29.9  & 14.0  & 55.0 & 0.2  & 15.1 & 57.4 & 23.3 & 15.2 & 56.0 & 23.5 & 11.8 & 82.1 & 69.6 & 43.4 & 48.1 & 60.4 \\
MiniCPM-o-2.6      & 85.3 & 0.7   & 0.0   & 42.8 & 66.9 & 29.8 & 54.5 & 35.4 & 12.3 & 25.1 & 17.2 & 46.6 & 62.7 & 45.7 & 42.8 & 41.3 & 49.5 \\
Gemma-3n-E4B       & 51.2 & 50.6 & 48.4 & 50.4 & 0.0 & 19.1 & 95.1 & 32.0 & 19.1 & 94.4 & 32.0 & 16.0 & 36.8 & 37.0 & 33.3 & 40.7 & 37.1 \\
Phi-4-multimodal   & 78.8 & 24.9 & 24.3 & 51.7 & 99.7 & 0.0 & 0.1 & 0.1 & 0.0 & 0.1 & 0.1 & 49.9 & 37.3 & 39.1 & 39.0 & 43.9 & 39.9 \\
Omnivinci          & 81.6 & 46.1  & 20.9  & 57.5 & 48.3 & 14.9 & 61.2 & 23.6 & 8.8  & 40.0 & 15.2 & 33.8 & 51.4 & 39.1 & 42.1 & 42.9 & 45.4 \\
Audio-Flamingo-3   & 99.2 & 1.3   & 1.2   & 50.2 & 0.0  & 17.1 & 38.4 & 23.3 & 17.4 & 38.8 & 23.7 & 11.8 & 33.0 & 28.3 & 37.1 & 41.3 & 36.3 \\
\bottomrule
\end{tabular}
\caption{Per-turn reference comparison vs.\ main 
results (multi-turn). In the per-turn setting, each turn is independently 
compared against the reference audio; the main setting presents 
all turns simultaneously. Detection reports per-scenario and 
Balanced Accuracy (Bal). Localization reports F1 for S1, 
Precision (P), Recall (R), and F1 for S2/S3, and Balanced F1. 
Discrimination reports per-dataset and average Classification 
Accuracy.}
\label{tab:pairwise_results}
\end{table*}

\section{Per-Turn Reference Comparison}
\label{app:pairwise}

Our main evaluation presents all target speaker turns 
simultaneously, allowing models to reason over the collective 
acoustic pattern across turns. Here we evaluate an alternative 
\textit{per-turn} setting that mirrors the Reference Comparison in speaker
embedding methods (Appendix~\ref{embedding_methods_details}): each turn 
is independently compared to the reference audio with the 
prompt ``Is this clip from the same speaker as the reference?'' 
(Figure~\ref{fig:prompt_perturn_det_loc}). For Detection, a dialogue is marked inconsistent if the model 
answers ``No'' on any turn; for Localization, all turns 
answered ``No'' are returned as the predicted inconsistent 
set. For Discrimination, each candidate is presented together 
with only the reference audio, omitting the surrounding 
consistent turns available in the main setting 
(Figure~\ref{fig:prompt_perturn_disc}).

Table~\ref{tab:pairwise_results} compares this setting to our 
main evaluation, revealing structural limitations across all 
three tasks. For Detection, most models collapse to flagging 
every dialogue: any single ``No'' response suffices to mark 
the dialogue inconsistent, yielding near-zero accuracy on S1 
and near-perfect on S2/S3. Balanced Detection Accuracy thus 
converges to $\sim$50\% across almost all models, compared to 
the 42.8--64.7\% spread in the main setting. Localization 
exhibits the same over-flagging: Recall exceeds 95\% for most 
models while Precision stays around 20\%. Discrimination also 
degrades---Gemini-2.5-Pro drops from 81.5\% to 50.3\% and 
Gemini-2.5-Flash from 75.6\% to 65.8\%---because a single 
reference clip provides a weaker acoustic anchor than the full 
set of consistent turns available in the main setting. These 
limitations underscore that multi-turn context is not merely a 
design complexity but a necessary condition for meaningful 
speaker consistency evaluation.

\begin{table*}[!ht]
\centering
\scriptsize
\setlength{\tabcolsep}{3pt}
\begin{tabular}{l|cccc|c|ccc|ccc|c|ccccc|ccc}
\toprule
& \multicolumn{4}{c|}{\textbf{Detection}} & \multicolumn{8}{c|}{\textbf{Localization}} & \multicolumn{8}{c}{\textbf{Discrimination}} \\
\cmidrule(lr){2-5} \cmidrule(lr){6-13} \cmidrule(lr){14-21}
\textbf{Model / Method} & S1 & S2 & S3 & Bal & \multicolumn{1}{c|}{S1} & \multicolumn{3}{c|}{S2} & \multicolumn{3}{c|}{S3} & Bal & \multicolumn{5}{c|}{Classification} & \multicolumn{3}{c}{Ranking} \\
\cmidrule(lr){2-5} \cmidrule(lr){6-6} \cmidrule(lr){7-9} \cmidrule(lr){10-12} \cmidrule(lr){13-13} \cmidrule(lr){14-18} \cmidrule(lr){19-21}
& \multicolumn{4}{c|}{Acc} & F1 & P & R & F1 & P & R & F1 & F1 & Baz & AMI & B-SD & Daily & Avg & N@1 & N@2 & EM \\
\midrule
\rowcolor{yellow!30}
\multicolumn{21}{c}{\textit{Large Audio-Language Models}} \\
GPT-4o-audio & 77.2 & 33.9 & 32.0 & 55.1 & 68.9 & 14.7 & 26.8 & 17.4 & 9.9 & 21.4 & 12.4 & 41.9 & 42.6 & 30.4 & 28.5 & 35.4 & 35.7 & 47.2 & 63.5 & 19.1 \\
Gemini-2.5-Pro & 66.2 & 65.1 & 43.9 & \underline{60.3} & 67.2 & 57.0 & 65.8 & 59.6 & 46.7 & 58.3 & 50.0 & \underline{61.0} & 86.7 & 82.6 & 82.9 & 91.5 & \textbf{86.9} & \textbf{93.5} & \textbf{93.9} & \textbf{72.8} \\
Gemini-2.5-Flash & 97.2 & 53.6 & 17.7 & \textbf{66.4} & 64.4 & 59.6 & 90.1 & 66.3 & 45.4 & 78.1 & 51.7 & \textbf{61.7} & 73.8 & 82.6 & 76.6 & 88.4 & \underline{79.8} & \underline{86.3} & \underline{90.1} & \underline{62.9} \\
Gemini-2.5-Flash-Lite & 32.7 & 75.4 & 74.3 & 53.8 & 0.7 & 23.6 & 94.7 & 36.2 & 23.2 & 92.5 & 35.4 & 18.2 & 25.7 & 50.0 & 45.0 & 51.1 & 40.6 & 56.5 & 68.0 & 25.0 \\
Qwen2.5-Omni-3B & 54.8 & 51.6 & 48.5 & 52.4 & 21.6 & 16.1 & 59.7 & 24.7 & 15.4 & 56.8 & 23.6 & 22.9 & 35.2 & 41.3 & 42.4 & 38.6 & 38.6 & 50.3 & 63.9 & 21.9 \\
Qwen2.5-Omni-7B & 32.8 & 70.9 & 70.5 & 51.8 & 38.3 & 14.5 & 64.0 & 23.2 & 13.3 & 58.6 & 21.2 & 30.3 & 44.3 & 37.0 & 34.2 & 40.7 & 40.0 & 50.6 & 64.0 & 22.2 \\
Qwen3-Omni-30B-A3B & 88.1 & 33.2 & 14.6 & 56.0 & 0.2 & 40.5 & 93.4 & 50.1 & 26.5 & 85.6 & 37.0 & 21.9 & 73.3 & 69.6 & 38.6 & 52.4 & 57.4 & 67.6 & 77.0 & 40.0 \\
MiniCPM-o-2.6 & 99.8 & 1.7 & 0.2 & 50.4 & 62.5 & 35.4 & 62.6 & 43.1 & 15.5 & 32.6 & 19.7 & 47.0 & 59.5 & 50.0 & 48.1 & 51.3 & 53.2 & 57.0 & 69.6 & 29.4 \\
Gemma-3n-E4B & 51.2 & 48.4 & 48.0 & 49.7 & 0.0 & 19.2 & 96.0 & 31.9 & 19.2 & 95.2 & 31.9 & 15.9 & 37.1 & 43.5 & 32.9 & 39.2 & 37.1 & 46.5 & 61.0 & 19.7 \\
Phi-4-multimodal & 79.0 & 24.1 & 23.3 & 51.4 & 99.7 & 0.0 & 0.0 & 0.0 & 0.1 & 0.3 & 0.1 & 49.9 & 36.2 & 39.1 & 41.1 & 45.0 & 40.5 & 47.6 & 62.8 & 20.6 \\
Omnivinci & 81.7 & 46.0 & 21.2 & 57.6 & 48.3 & 14.5 & 63.3 & 23.3 & 9.5 & 43.9 & 15.5 & 33.9 & 41.9 & 50.0 & 41.8 & 50.8 & 45.3 & 56.2 & 67.7 & 27.4 \\
Audio-Flamingo-3 & 99.2 & 1.3 & 1.0 & 50.2 & 0.0 & 17.2 & 38.7 & 23.5 & 17.3 & 38.7 & 23.6 & 11.8 & 32.9 & 28.3 & 36.1 & 42.9 & 36.5 & 44.3 & 61.2 & 16.9 \\
\midrule
\rowcolor{blue!20}
\multicolumn{21}{c}{\textit{Speaker Embedding Methods}} \\
Pairwise (ECAPA) & 26.9 & 98.2 & 94.2 & 61.5 & 26.9 & 42.8 & 92.9 & 53.8 & 37.5 & 82.1 & 47.0 & 38.7 & 99.5 & 97.8 & 99.4 & 100.0 & \textbf{99.5} & \textbf{99.6} & \textbf{93.2} & \textbf{61.2} \\
Pairwise (WavLM) & 96.7 & 35.7 & 36.3 & \underline{66.3} & 96.7 & 32.6 & 33.6 & 32.9 & 33.5 & 34.3 & 33.8 & \underline{65.0} & 92.9 & 97.8 & 81.0 & 98.9 & \underline{92.0} & \underline{94.2} & \underline{91.5} & \underline{55.7} \\
Centroid (ECAPA) & 21.6 & 99.5 & 96.0 & 59.7 & 21.6 & 52.6 & 91.1 & 62.8 & 46.7 & 81.1 & 55.8 & 40.5 & 99.5 & 97.8 & 99.4 & 100.0 & \textbf{99.5} & \textbf{99.6} & \textbf{93.2} & \textbf{61.2} \\
Centroid (WavLM) & 91.4 & 53.4 & 56.3 & \textbf{73.1} & 91.4 & 48.5 & 49.9 & 49.0 & 51.0 & 52.3 & 51.4 & \textbf{70.8} & 92.9 & 97.8 & 81.0 & 98.9 & \underline{92.0} & \underline{94.2} & \underline{91.5} & \underline{55.7} \\
Reference (ECAPA) & 5.8 & 99.0 & 96.5 & 51.8 & 5.8 & 29.7 & 90.1 & 41.9 & 25.0 & 81.1 & 36.1 & 22.4 & 93.3 & 97.8 & 99.4 & 78.4 & 90.6 & 91.3 & 86.4 & 53.1 \\
Reference (WavLM) & 91.1 & 27.6 & 27.5 & 59.3 & 91.1 & 20.4 & 21.5 & 20.7 & 19.1 & 20.0 & 19.4 & 55.7 & 92.4 & 97.8 & 74.1 & 70.8 & 81.3 & 86.5 & 81.2 & 49.5 \\
\bottomrule
\end{tabular}
\caption{Full evaluation results on the extended benchmark, 
which applies four voice conversion models to the original 
dialogues (FreeVC, CosyVoice3, OpenVoice, YourTTS) (\%). 
Detection reports per-scenario and Balanced Accuracy (Bal). 
Localization reports F1 for S1, Precision (P), Recall (R), 
and F1 for S2/S3, and Balanced F1. Discrimination reports 
per-dataset and average Classification Accuracy, and Ranking 
metrics (N@1: NDCG@1, N@2: NDCG@2, EM: Exact Match).}
\label{tab:extended_vc_full}
\end{table*}

\section{Extended Benchmarks}
\label{app:extended}

\subsection{Multiple Voice Conversion Models}
\label{app:extended_vc}

To verify that our benchmark findings are not specific to a 
single voice conversion (VC) model, we construct an extended 
benchmark by applying three additional VC models---CosyVoice3~\cite{du2025cosyvoice3}, 
OpenVoice~\cite{qin2023openvoice}, and YourTTS~\cite{casanova2022yourtts}---alongside 
the original FreeVC~\cite{li2023freevc}. The extended benchmark 
shares exactly the same set of multi-turn dialogue instances as 
the original: the same dialogues, target speakers, and scenario 
structure, with the only difference being which VC model 
generates the inconsistent audio. This design enables a direct, 
controlled comparison isolating the effect of VC model 
variability from all other factors. The resulting inconsistent 
clips are distributed as follows: CosyVoice3 (44\%), OpenVoice 
(24\%), FreeVC (23\%), and YourTTS (9\%).

Table~\ref{tab:extended_vc_full} reports the complete evaluation 
results on the extended benchmark, following the same format as 
our main results (Table~\ref{tab:main_results}). 
Table~\ref{tab:extended_vc} summarizes the comparison against 
the original FreeVC-only benchmark using the three headline 
metrics: Balanced Detection Accuracy, Balanced Localization 
F1-score, and Discrimination Accuracy. The performance 
differences are small for most models: the average absolute 
$\Delta$ is 1.9\% for Detection, 2.0\% for Localization, and 
2.8\% for Discrimination, with no systematic direction of 
change. Model rankings are generally preserved, and the key qualitative 
patterns from Section~\ref{sec:main_results} hold under the 
extended benchmark: the anti-diagonal clustering indicating 
threshold instability, the Precision/Recall divergence in 
Localization, and the detection--discrimination dissociation.

\begin{table}[h]
\centering
\resizebox{\linewidth}{!}{%
\begin{tabular}{l|ccc|ccc|ccc}
\toprule
\multirow{2}{*}{Model} & \multicolumn{3}{c|}{Detection} & \multicolumn{3}{c|}{Localization} & \multicolumn{3}{c}{Discrimination} \\
& FVC & 4VC & $\Delta$ & FVC & 4VC & $\Delta$ & FVC & 4VC & $\Delta$ \\
\midrule
GPT-4o-audio & 52.0 & 55.1 & \inc{3.1} & 42.3 & 41.9 & \dec{0.4} & 40.7 & 35.7 & \dec{5.0} \\
Gemini-2.5-Pro & \textbf{64.7} & 60.3 & \dec{4.4} & \textbf{60.1} & 61.0 & \inc{0.9} & \textbf{81.5} & \textbf{86.9} & \inc{5.4} \\
Gemini-2.5-Flash & 63.1 & \textbf{66.4} & \inc{3.3} & 59.0 & \textbf{61.7} & \inc{2.7} & 75.6 & 79.8 & \inc{4.2} \\
Gemini-2.5-Flash-Lite & 55.3 & 53.8 & \dec{1.5} & 18.1 & 18.2 & \inc{0.1} & 46.5 & 40.6 & \dec{5.9} \\
Qwen2.5-Omni-3B & 51.5 & 52.4 & \inc{0.9} & 31.8 & 22.9 & \dec{8.9} & 40.8 & 38.6 & \dec{2.2} \\
Qwen2.5-Omni-7B & 51.5 & 51.8 & \inc{0.3} & 29.9 & 30.3 & \inc{0.4} & 42.7 & 40.0 & \dec{2.7} \\
Qwen3-Omni-30B-A3B & 55.0 & 56.0 & \inc{1.0} & 11.8 & 21.9 & \inc{10.1} & 60.4 & 57.4 & \dec{3.0} \\
MiniCPM-o-2.6 & 42.8 & 50.4 & \inc{7.6} & 46.6 & 47.0 & \inc{0.4} & 49.5 & 53.2 & \inc{3.7} \\
Gemma-3n-E4B & 50.4 & 49.7 & \dec{0.7} & 16.0 & 15.9 & \dec{0.1} & 37.1 & 37.1 & 0.0 \\
Phi-4-multimodal & 51.7 & 51.4 & \dec{0.3} & 49.9 & 49.9 & 0.0 & 39.9 & 40.5 & \inc{0.6} \\
Omnivinci & 57.5 & 57.6 & \inc{0.1} & 33.8 & 33.9 & \inc{0.1} & 45.4 & 45.3 & \dec{0.1} \\
Audio-Flamingo-3 & 50.2 & 50.2 & 0.0 & 11.8 & 11.8 & 0.0 & 36.3 & 36.5 & \inc{0.2} \\
\bottomrule
\end{tabular}%
}
\caption{Comparison of model performance on the original 
benchmark (FVC: FreeVC only) versus the extended benchmark 
(4VC: FreeVC + CosyVoice3 + OpenVoice + YourTTS). Det: 
Balanced Accuracy. Loc: Balanced
F1. Disc: Avg Classification 
Accuracy.}
\label{tab:extended_vc}
\vspace{-1em}
\end{table}

\subsection{Longer Dialogues}
\label{app:10turn}

\begin{table*}[!ht]
\centering
\scriptsize
\setlength{\tabcolsep}{3pt}
\begin{tabular}{l|cccc|c|ccc|ccc|c|ccccc|ccc}
\toprule
& \multicolumn{4}{c|}{\textbf{Detection}} & \multicolumn{8}{c|}{\textbf{Localization}} & \multicolumn{8}{c}{\textbf{Discrimination}} \\
\cmidrule(lr){2-5} \cmidrule(lr){6-13} \cmidrule(lr){14-21}
\textbf{Model / Method} & S1 & S2 & S3 & Bal & \multicolumn{1}{c|}{S1} & \multicolumn{3}{c|}{S2} & \multicolumn{3}{c|}{S3} & Bal & \multicolumn{5}{c|}{Classification} & \multicolumn{3}{c}{Ranking} \\
\cmidrule(lr){2-5} \cmidrule(lr){6-6} \cmidrule(lr){7-9} \cmidrule(lr){10-12} \cmidrule(lr){13-13} \cmidrule(lr){14-18} \cmidrule(lr){19-21}
& \multicolumn{4}{c|}{Acc} & F1 & P & R & F1 & P & R & F1 & F1 & Baz & AMI & B-SD & Daily & Avg & N@1 & N@2 & EM \\
\midrule
\rowcolor{yellow!30}
\multicolumn{21}{c}{\textit{Large Audio-Language Models}} \\
GPT-4o-audio & 85.4 & 18.6 & 15.8 & 51.3 & 70.4 & 7.0 & 13.4 & 8.8 & 4.3 & 10.7 & 5.8 & 38.8 & 60.0 & 41.4 & 35.4 & 52.4 & 39.2 & 45.6 & 60.5 & 12.0 \\
Gemini-2.5-Pro & 73.5 & 75.9 & 54.2 & \textbf{69.3} & 63.6 & 46.3 & 60.1 & 49.9 & 40.2 & 56.1 & 44.4 & \underline{55.4} & 80.0 & 71.2 & 81.4 & 95.2 & \underline{80.1} & \textbf{87.7} & \textbf{90.0} & \textbf{63.1} \\
Gemini-2.5-Flash & 100.0 & 22.5 & 2.8 & \underline{56.3} & 74.7 & 59.1 & 83.4 & 63.4 & 33.1 & 62.8 & 36.8 & \textbf{62.4} & 72.7 & 81.7 & 81.4 & 95.2 & \textbf{82.2} & \underline{85.5} & \underline{88.8} & \underline{61.3} \\
Gemini-2.5-Flash-Lite & 43.1 & 51.4 & 54.9 & 48.1 & 12.6 & 10.9 & 74.7 & 17.1 & 11.1 & 84.2 & 18.1 & 15.1 & 41.1 & 55.6 & 47.4 & 52.9 & 46.6 & 49.9 & 59.4 & 19.4 \\
Qwen2.5-Omni-3B & 76.3 & 28.1 & 26.5 & 51.8 & 29.2 & 6.2 & 52.2 & 10.6 & 5.9 & 52.6 & 10.2 & 19.8 & 54.5 & 43.3 & 39.8 & 33.3 & 40.7 & 46.4 & 60.2 & 19.4 \\
Qwen2.5-Omni-7B & 48.2 & 54.2 & 53.8 & 51.1 & 64.0 & 5.1 & 29.6 & 8.0 & 3.8 & 25.7 & 6.3 & 35.6 & 81.8 & 48.3 & 31.7 & 42.9 & 38.7 & 48.1 & 61.5 & 20.2 \\
Qwen3-Omni-30B-A3B & 97.6 & 5.5 & 2.8 & 50.9 & 1.2 & 30.4 & 84.6 & 36.1 & 16.7 & 77.9 & 23.1 & 15.4 & 72.7 & 68.3 & 48.4 & 57.1 & 54.9 & 59.4 & 68.0 & 30.0 \\
MiniCPM-o-2.6 & 100.0 & 0.4 & 0.4 & 50.2 & 51.0 & 22.7 & 45.8 & 29.2 & 10.8 & 24.5 & 14.2 & 36.3 & 54.5 & 51.7 & 30.4 & 47.6 & 37.9 & 48.5 & 63.8 & 20.2 \\
Gemma-3n-E4B & 72.7 & 30.8 & 29.6 & 51.5 & 0.0 & 8.9 & 94.5 & 16.3 & 9.0 & 95.7 & 16.5 & 8.2 & 18.2 & 36.7 & 36.6 & 28.6 & 35.2 & 44.9 & 59.9 & 18.6 \\
Phi-4-multimodal & 95.3 & 5.9 & 4.3 & 50.2 & 100.0 & 0.0 & 0.0 & 0.0 & 0.0 & 0.0 & 0.0 & 50.0 & 54.5 & 28.3 & 29.2 & 47.6 & 31.6 & 44.3 & 59.7 & 18.6 \\
Omnivinci & 79.8 & 27.3 & 20.9 & 52.0 & 22.5 & 6.4 & 66.4 & 11.4 & 5.0 & 58.1 & 9.2 & 16.4 & 36.4 & 41.7 & 39.8 & 42.9 & 40.3 & 51.4 & 64.5 & 21.3 \\
Audio-Flamingo-3 & 92.9 & 4.7 & 3.6 & 48.5 & 0.4 & 9.5 & 56.5 & 15.2 & 9.3 & 58.1 & 15.0 & 7.7 & 54.5 & 28.3 & 29.2 & 52.4 & 32.0 & 43.7 & 60.2 & 17.0 \\
\midrule
\end{tabular}
\caption{Full evaluation results on the 10-turn benchmark (\%). Detection reports per-scenario and Balanced Accuracy (Bal). Localization reports F1 for S1, Precision (P), Recall (R), and F1 for S2/S3, and Balanced F1. Discrimination reports per-dataset and average Classification Accuracy, and Ranking metrics (N@1: NDCG@1, N@2: NDCG@2, EM: Exact Match).}
\label{tab:10turn_full}
\end{table*}

To examine whether our findings generalize across dialogue 
lengths, we construct a 10-turn benchmark comprising 759 
instances from 253 unique dialogues using the same pipeline 
as our primary 5-turn benchmark 
(Section~\ref{sec:speakersleuth}). This extension preserves all 
benchmark construction components and changes only the number of target speaker 
turns.

Table~\ref{tab:10turn_full} reports the full per-scenario 
Detection, Localization, and Discrimination results on the 
10-turn benchmark, following the same format as our main 
results (Table~\ref{tab:main_results}). 
Section~\ref{sec:10turn} summarizes these results 
using the three headline metrics 
(Table~\ref{tab:10turn}) and discusses the overall trend: 
the 10-turn setting is slightly more challenging on average 
($-1.2\%$ Detection, $-4.2\%$ Localization, $-3.1\%$ 
Discrimination), while model rankings are largely preserved. The full per-scenario breakdowns here confirm that the key 
qualitative patterns observed in the 5-turn setting remain 
intact. The anti-diagonal clustering persists 
(e.g., Gemini-2.5-Flash reaches 100\% on S1 but only 
22.5\%/2.8\% on S2/S3, while Gemini-2.5-Flash-Lite shows 
the opposite bias), Precision/Recall divergence in 
Localization continues (Gemma-3n-E4B: P~$\approx$~9\%, 
R~$\approx$~95\% on S2), and Gemini-2.5-Pro remains strong 
on Discrimination (80.1\% Avg, 63.1\% Exact Match) despite 
its modest Detection performance (69.3\% Balanced Accuracy), 
confirming the detection--discrimination dissociation.

\begin{table*}[h]
\centering
\scriptsize
\setlength{\tabcolsep}{2.8pt}
\begin{tabular}{l|rrr|rrr|rrr||rrr|rrr|rrr}
\toprule
& \multicolumn{9}{c||}{\textbf{Impact of Textual Context}} & \multicolumn{9}{c}{\textbf{Impact of Reference Audio}} \\
\cmidrule(lr){2-10} \cmidrule(lr){11-19}
& \multicolumn{3}{c|}{\textbf{S1}} & \multicolumn{3}{c|}{\textbf{S2}} & \multicolumn{3}{c||}{\textbf{S3}} & \multicolumn{3}{c|}{\textbf{S1}} & \multicolumn{3}{c|}{\textbf{S2}} & \multicolumn{3}{c}{\textbf{S3}} \\
\cmidrule(lr){2-4} \cmidrule(lr){5-7} \cmidrule(lr){8-10} \cmidrule(lr){11-13} \cmidrule(lr){14-16} \cmidrule(lr){17-19}
\textbf{Model} & Audio & +C & $\Delta$ & Audio & +C & $\Delta$ & Audio & +C & $\Delta$ & w/ Ref & w/o & $\Delta$ & w/ Ref & w/o & $\Delta$ & w/ Ref & w/o & $\Delta$ \\
\midrule
GPT-4o-audio & 71.5 & 88.6 & \inc{17.1} & 15.0 & 3.6 & \dec{11.4} & 11.1 & 3.7 & \dec{7.4} & 71.5 & 79.9 & \inc{8.4} & 15.0 & 11.4 & \dec{3.6} & 11.1 & 7.2 & \dec{3.9} \\
Gemini-2.5-Pro & 65.2 & 97.3 & \inc{32.1} & 62.5 & 7.9 & \dec{54.6} & 47.5 & 6.3 & \dec{41.2} & 65.2 & 87.3 & \inc{22.1} & 62.5 & 44.0 & \dec{18.5} & 47.5 & 28.2 & \dec{19.3} \\
Gemini-2.5-Flash & 64.4 & 98.8 & \inc{34.4} & 62.3 & 30.3 & \dec{32.0} & 45.0 & 14.3 & \dec{30.7} & 64.4 & 94.1 & \inc{29.7} & 62.3 & 57.9 & \dec{4.4} & 45.0 & 32.4 & \dec{12.6} \\
Gemini-2.5-Flash-Lite & 0.3 & 15.0 & \inc{14.7} & 36.4 & 29.6 & \dec{6.8} & 35.4 & 31.0 & \dec{4.4} & 0.3 & 6.6 & \inc{6.3} & 36.4 & 42.5 & \inc{6.1} & 35.4 & 41.0 & \inc{5.6} \\
Qwen2.5-Omni-3B & 21.8 & 76.0 & \inc{54.2} & 47.7 & 3.7 & \dec{44.0} & 36.0 & 4.1 & \dec{31.9} & 21.8 & 82.8 & \inc{61.0} & 47.7 & 4.7 & \dec{43.0} & 36.0 & 4.2 & \dec{31.8} \\
Qwen2.5-Omni-7B & 38.0 & 97.8 & \inc{59.8} & 22.6 & 0.6 & \dec{22.0} & 21.0 & 0.3 & \dec{20.7} & 38.0 & 88.3 & \inc{50.3} & 22.6 & 3.2 & \dec{19.4} & 21.0 & 2.8 & \dec{18.2} \\
Qwen3-Omni-30B-A3B & 0.2 & 26.4 & \inc{26.2} & 23.3 & 36.3 & \inc{13.0} & 23.5 & 26.0 & \inc{2.5} & 0.2 & 1.3 & \inc{1.1} & 23.3 & 52.6 & \inc{29.3} & 23.5 & 30.4 & \inc{6.9} \\
MiniCPM-o-2.6 & 66.9 & 59.0 & \dec{7.9} & 35.4 & 23.9 & \dec{11.5} & 17.2 & 14.9 & \dec{2.3} & 66.9 & 62.8 & \dec{4.1} & 35.4 & 38.9 & \inc{3.5} & 17.2 & 19.7 & \inc{2.5} \\
Gemma-3n-E4B & 0.0 & 1.3 & \inc{1.3} & 32.0 & 28.0 & \dec{4.0} & 32.0 & 27.9 & \dec{4.1} & 0.0 & 0.0 & \neu{0.0} & 32.0 & 30.7 & \dec{1.3} & 32.0 & 30.2 & \dec{1.8} \\
Phi-4-multimodal & 99.7 & 100.0 & \inc{0.3} & 0.1 & 0.0 & \dec{0.1} & 0.1 & 0.0 & \dec{0.1} & 99.7 & 100.0 & \inc{0.3} & 0.1 & 0.0 & \dec{0.1} & 0.1 & 0.0 & \dec{0.1} \\
Omnivinci & 48.3 & 17.8 & \dec{30.5} & 23.6 & 10.7 & \dec{12.9} & 15.2 & 9.8 & \dec{5.4} & 48.3 & 90.0 & \inc{41.7} & 23.6 & 8.9 & \dec{14.7} & 15.2 & 2.7 & \dec{12.5} \\
Audio-Flamingo-3 & 0.0 & 3.5 & \inc{3.5} & 23.3 & 17.6 & \dec{5.7} & 23.7 & 17.8 & \dec{5.9} & 0.0 & 0.7 & \inc{0.7} & 23.3 & 23.3 & \neu{0.0} & 23.7 & 22.7 & \dec{1.0} \\
\bottomrule
\end{tabular}
\caption{\textbf{Impact of Textual Context and Reference Audio on Localization F1 (\%).} Left: performance change when adding textual context (+C) vs.\ audio-only (Audio). Right: performance change when removing reference audio (w/o) vs.\ with reference (w/ Ref). $\Delta$ denotes the difference.}
\label{tab:analysis_impact_localization}
\end{table*}

\section{Impact of Textual Context and Reference Audio on Localization}
\label{app:localization_ablations}

Table~\ref{tab:analysis_impact_localization} reports the 
Localization F1 results for the same ablation conditions 
analyzed for Detection in Section~\ref{sec:analysis}: adding 
textual context (+C vs.\ Audio) and removing reference audio 
(w/o vs.\ w/ Ref).

Both conditions reproduce the asymmetric behavior observed for 
Detection. Adding textual context sharply improves S1 F1 
(e.g., Qwen2.5-Omni-7B: 38.0~$\rightarrow$~97.8) while 
collapsing S2 and S3 F1 (e.g., Gemini-2.5-Pro drops by 
$-54.6\%$ on S2 and $-41.2\%$ on S3), confirming that textual 
context biases models toward declaring all target-speaker 
turns as consistent, even when an inconsistent turn is 
present. Removing reference audio produces an analogous 
pattern for several models (e.g., Qwen2.5-Omni-3B S2: 
47.7~$\rightarrow$~4.7, S3: 36.0~$\rightarrow$~4.2), though 
the direction and magnitude of change are more variable 
across models than in the textual context setting. These 
results mirror the Detection findings in 
Section~\ref{sec:analysis_textual} 
and~\ref{sec:analysis_reference}, demonstrating that the 
observed modality imbalances are not specific to the 
Detection formulation but extend to fine-grained turn-level 
localization.

\section{Potential Risks}
Despite positive applications, we acknowledge several risks. Advanced speaker consistency evaluation could be exploited to create more convincing synthetic voices for impersonation, fraud, or misinformation campaigns. As synthetic speech becomes increasingly indistinguishable from human speech, public trust in audio evidence and voice-based authentication may diminish. Additionally, technologies that analyze speaker characteristics could be misappropriated for unauthorized surveillance or profiling.

\section{AI Assistants in Research or Writing}
We used AI assistants to refine and proofread the text, and 
assist with coding experiments. However, all core ideas, 
experimental design, analysis, and scientific contributions 
are entirely the work of the authors.

\section{Prompt Templates}
\label{appendix:prompt}
\begin{figure*}[t]
\centering
\begin{tcolorbox}[title={Prompt Template for Detection}, halign=left, boxrule=0.5pt]\normalsize
You are an expert at speaker recognition. You can determine if audio turns are consistent with a target speaker based on voice characteristics alone. Focus on the main speaker's voice characteristics. Ignore backchannels, background noise, or short interjections.
\vspace{2ex}

First, listen to this reference audio clip from the target speaker:
\vspace{1ex}

[Audio: \texttt{\{reference\_audio\}}]
\vspace{2ex}

Now listen to the following turns from the conversation. Focus ONLY on the voice identity.
\vspace{2ex}

Turn 1: [Audio: \texttt{\{turn\_1\}}] \\
Turn 2: [Audio: \texttt{\{turn\_2\}}] \\
\dots \\
Turn $N$: [Audio: \texttt{\{turn\_N\}}]
\vspace{2ex}

Question: Is the voice consistent across all these turns?
\vspace{2ex}

Answer with ONLY one word: YES or NO.
\end{tcolorbox}
\caption{Prompt template for Detection: determining speaker consistency across dialogue turns using audio only.}
\label{fig:prompt_detection_audio_only}
\end{figure*}

\begin{figure*}[h]
\centering
\begin{tcolorbox}[title={Prompt Template for Detection (with Textual Context)}, halign=left, boxrule=0.5pt]\normalsize
You are an expert at speaker recognition. You can determine if audio turns provided for a specific speaker in a conversation are consistent. Focus on the main speaker's voice characteristics. Ignore backchannels, background noise, or short interjections.
\vspace{2ex}

First, listen to this reference audio clip from the target speaker (\texttt{\{target\_speaker\}}):
\vspace{1ex}

[Audio: \texttt{\{reference\_audio\}}]
\vspace{2ex}

Now listen to the following conversation. Focus on the turns spoken by \texttt{\{target\_speaker\}}.
\vspace{2ex}

Turn 1 (\texttt{\{speaker\_1\}}): [Audio: \texttt{\{turn\_1\}}] \\
Turn 2 (\texttt{\{speaker\_2\}}): \texttt{\{text\_2\}} \\
\dots \\
Turn $N$ (\texttt{\{speaker\_N\}}): [Audio: \texttt{\{turn\_N\}}]
\vspace{2ex}

Question: Is the voice of \texttt{\{target\_speaker\}} consistent across all their turns in this conversation?
\vspace{2ex}

Answer with ONLY one word: YES or NO.
\end{tcolorbox}
\caption{Prompt template for Detection (with Textual Context): determining speaker consistency when textual transcripts of other speakers' turns are also provided.}
\label{fig:prompt_detection_with_text}
\end{figure*}


\begin{figure*}[h]
\centering
\begin{tcolorbox}[title={Prompt Template for Localization}, halign=left, boxrule=0.5pt]\normalsize
You are an expert at speaker recognition. You can identify if audio turns are from the same speaker by analyzing voice characteristics alone. Focus on the main speaker's voice characteristics. Ignore backchannels, background noise, or short interjections.
\vspace{2ex}

First, listen to this reference audio clip from the target speaker:
\vspace{1ex}

[Audio: \texttt{\{reference\_audio\}}]
\vspace{2ex}

Now listen to the following turns. Identify which turns (if any) are likely spoken by a different speaker than the target speaker.
\vspace{2ex}

Turn 1: [Audio: \texttt{\{turn\_1\}}] \\
Turn 2: [Audio: \texttt{\{turn\_2\}}] \\
\dots \\
Turn $N$: [Audio: \texttt{\{turn\_N\}}]
\vspace{2ex}

Question: Which turns are inconsistent with the target 
speaker? List the turn numbers (e.g., ``Turn 3, Turn 5''). 
If all turns are consistent, answer ``None''.
\end{tcolorbox}
\caption{Prompt template for Localization: identifying which turns are inconsistent with the target speaker using audio only.}
\label{fig:prompt_localization_audio_only}
\end{figure*}

\begin{figure*}[h]
\centering
\begin{tcolorbox}[title={Prompt Template for Localization (with Textual Context)}, halign=left, boxrule=0.5pt]\normalsize
You are an expert at speaker recognition. You can determine if audio turns provided for a specific speaker in a conversation are consistent. Focus on the main speaker's voice characteristics. Ignore backchannels, background noise, or short interjections.
\vspace{2ex}

First, listen to this reference audio clip from the target speaker (\texttt{\{target\_speaker\}}):
\vspace{1ex}

[Audio: \texttt{\{reference\_audio\}}]
\vspace{2ex}

Now listen to the following conversation involving \texttt{\{target\_speaker\}}. Identify which turns (if any) attributed to \texttt{\{target\_speaker\}} are likely spoken by a different speaker than the target speaker.
\vspace{2ex}

Turn 1 (\texttt{\{speaker\_1\}}): [Audio: \texttt{\{turn\_1\}}] \\
Turn 2 (\texttt{\{speaker\_2\}}): \texttt{\{text\_2\}} \\
\dots \\
Turn $N$ (\texttt{\{speaker\_N\}}): [Audio: \texttt{\{turn\_N\}}]
\vspace{2ex}

Question: Which turns are inconsistent with the target 
speaker? List the turn numbers (e.g., ``Turn 3, Turn 5''). 
If all turns are consistent, answer ``None''.
\end{tcolorbox}
\caption{Prompt template for Localization (with Textual Context): identifying which turns are inconsistent when textual transcripts of other speakers' turns are also provided.}
\label{fig:prompt_localization_with_text}
\end{figure*}


\begin{figure*}[h]
\centering
\begin{tcolorbox}[title={Prompt Template for Discrimination (Classification)}, halign=left, boxrule=0.5pt]\normalsize
You are an expert at speaker recognition. You can identify speaker consistency by analyzing voice characteristics alone. Focus on the main speaker's voice characteristics. Ignore backchannels, background noise, or short interjections.
\vspace{2ex}

You will hear a sequence of turns. One position in the sequence is MISSING. You will be given three options for what should go in that position. Your task is to select the option that makes the entire sequence most consistent with the target speaker.
\vspace{2ex}

Listen to the turn sequence:
\vspace{1ex}

Turn 1: [Audio: \texttt{\{turn\_1\}}] \\
\dots \\
Turn $k$: [MISSING -- Choose the correct option below] \\
\dots \\
Turn $N$: [Audio: \texttt{\{turn\_N\}}]
\vspace{2ex}

Here are the options for the missing position (Turn $k$):
\vspace{1ex}

Option A: [Audio: \texttt{\{option\_A\}}] \\
Option B: [Audio: \texttt{\{option\_B\}}] \\
Option C: [Audio: \texttt{\{option\_C\}}]
\vspace{2ex}

Question: Which option (A, B, or C) makes the entire sequence most consistent with the target speaker?
\vspace{2ex}

Answer with ONLY one letter: A, B, or C.
\end{tcolorbox}
\caption{Prompt template for Discrimination (Classification): selecting the option that best fits the missing position to maximize speaker consistency.}
\label{fig:prompt_discrimination_classification}
\end{figure*}

\begin{figure*}[h]
\centering
\begin{tcolorbox}[title={Prompt Template for Discrimination (Ranking)}, halign=left, boxrule=0.5pt]\normalsize
You are an expert at speaker recognition. You can identify speaker consistency by analyzing voice characteristics alone. Focus on the main speaker's voice characteristics. Ignore backchannels, background noise, or short interjections.
\vspace{2ex}

You will hear a sequence of turns. One position in the sequence is MISSING. You will be given three options for what should go in that position. Your task is to rank all options by how well they match the target speaker's voice.
\vspace{2ex}

Listen to the turn sequence:
\vspace{1ex}

Turn 1: [Audio: \texttt{\{turn\_1\}}] \\
\dots \\
Turn $k$: [MISSING -- Rank the options below] \\
\dots \\
Turn $N$: [Audio: \texttt{\{turn\_N\}}]
\vspace{2ex}

Here are the options for the missing position (Turn $k$):
\vspace{1ex}

Option A: [Audio: \texttt{\{option\_A\}}] \\
Option B: [Audio: \texttt{\{option\_B\}}] \\
Option C: [Audio: \texttt{\{option\_C\}}]
\vspace{2ex}

Question: Rank all options (A, B, C) from most consistent to least consistent with the target speaker.
\vspace{2ex}

Answer with ONLY the letters in order, separated by ``>''. For example: A > B > C.
\end{tcolorbox}
\caption{Prompt template for Discrimination (Ranking): ranking candidates by acoustic consistency with the target speaker given the surrounding dialogue context.}
\label{fig:prompt_discrimination_ranking}
\end{figure*}


\begin{figure*}[h]
\centering
\begin{tcolorbox}[title={Prompt Template for Per-Turn Reference Comparison (Detection and Localization)}, halign=left, boxrule=0.5pt]\normalsize
You are an expert at speaker recognition. You can determine if two audio clips are from the same speaker by analyzing voice characteristics alone. Focus on the main speaker's voice characteristics. Ignore backchannels, background noise, or short interjections.
\vspace{2ex}

First, listen to this reference audio clip from the target speaker:
\vspace{1ex}

[Audio: \texttt{\{reference\_audio\}}]
\vspace{2ex}

Now listen to this audio clip:
\vspace{1ex}

[Audio: \texttt{\{test\_audio\}}]
\vspace{2ex}

Question: Is this clip from the same speaker as the reference?
\vspace{2ex}

Answer with ONLY one word: YES or NO.
\end{tcolorbox}
\caption{Prompt template for Per-Turn Reference Comparison (Detection and Localization): determining whether each target speaker turn is from the same speaker as the reference.}
\label{fig:prompt_perturn_det_loc}
\end{figure*}

\begin{figure*}[h]
\centering
\begin{tcolorbox}[title={Prompt Template for Per-Turn Reference Comparison (Discrimination)}, halign=left, boxrule=0.5pt]\normalsize
You are an expert at speaker recognition. You can determine the speaker's identity in audio recordings by analyzing their voice characteristics. Focus on the main speaker's voice characteristics. Ignore backchannels, background noise, or short interjections.
\vspace{2ex}

First, listen to this reference audio clip from the target speaker:
\vspace{1ex}

[Audio: \texttt{\{reference\_audio\}}]
\vspace{2ex}

You will hear three versions of the same utterance. Your 
task is to select the option that is most consistent with 
the target speaker.

\vspace{2ex}

Option A: [Audio: \texttt{\{option\_A\}}] \\
Option B: [Audio: \texttt{\{option\_B\}}] \\
Option C: [Audio: \texttt{\{option\_C\}}]
\vspace{2ex}

Question: Which option (A, B, or C) is most consistent with the target speaker?
\vspace{2ex}

Answer with ONLY one letter: A, B, or C.
\end{tcolorbox}
\caption{Prompt template for Per-Turn Reference Comparison (Discrimination): selecting the original target speaker audio among three candidates compared against the reference.}
\label{fig:prompt_perturn_disc}
\end{figure*}


\begin{figure*}[h]
\centering
\begin{tcolorbox}[title={Prompt Template for VC Quality Ranking}, halign=left, boxrule=0.5pt]\normalsize
You are an expert at speaker recognition. You can identify speakers by analyzing voice characteristics such as pitch, tone, speaking style, and timbre. Focus on the main speaker's voice characteristics. Ignore backchannels, background noise, or short interjections.
\vspace{2ex}

First, listen to this reference audio clip from the target speaker:
\vspace{1ex}

[Audio: \texttt{\{reference\_audio\}}]
\vspace{2ex}

Now, you will hear several audio clips. Your task is to rank all clips by how similar they sound to the reference speaker's voice.
\vspace{2ex}

Option A: [Audio: \texttt{\{option\_A\}}] \\
Option B: [Audio: \texttt{\{option\_B\}}] \\
Option C: [Audio: \texttt{\{option\_C\}}] \\
Option D: [Audio: \texttt{\{option\_D\}}]
\vspace{2ex}

Question: Rank all options (A, B, C, D) from most similar to least similar to the reference speaker's voice.
\vspace{2ex}

Answer with ONLY the letters in order, separated by ``>''. For example: A > B > C > D.
\end{tcolorbox}
\caption{Prompt template for VC Quality Ranking: ranking voice-cloned candidates by acoustic similarity to the reference speaker.}
\label{fig:prompt_vc_ranking}
\end{figure*}


\begin{figure*}[h]
\centering
\begin{tcolorbox}[title={Prompt Template for Dialogue Coherence Filtering}, halign=left, boxrule=0.5pt]\normalsize
You are an expert dialogue analyzer. Read the following conversation and determine if it flows naturally and coherently as a whole.
\vspace{2ex}

Conversation:
\vspace{1ex}

\texttt{\{speaker\_1\}}: \texttt{\{text\_1\}} \\
\texttt{\{speaker\_2\}}: \texttt{\{text\_2\}} \\
\dots \\
\texttt{\{speaker\_N\}}: \texttt{\{text\_N\}}
\vspace{2ex}

Task:
\vspace{1ex}

If the conversation flows naturally with logical connections between turns, output ``coherent''. If there are sudden disruptions, random insertions, contradictions, or parts that don't make sense in context (at any point in the dialogue), output ``incoherent''.
\vspace{2ex}

Return ONLY one word: ``coherent'' or ``incoherent''.
\end{tcolorbox}
\caption{Prompt template for Dialogue Coherence Filtering: evaluating conversational naturalness and flow for benchmark construction.}
\label{fig:prompt_coherence_filter}
\end{figure*}
\end{document}